\pdfoutput=1
\newif\ifarxiv
\arxivtrue
\newif\ifafour
\afourfalse% true = A4, false = A5
\newif\iftypodisclaim % typographical disclaim on the side
\typodisclaimfalse

\ifafour
\documentclass[a4paper,12pt,%extrafontsizes,%
onecolumn,oneside,article,%french,italian,german,swedish,latin,
british%
]{memoir}
\else
\documentclass[a5paper,10pt,%extrafontsizes,%
onecolumn,oneside,article,%french,italian,german,swedish,latin,
british%
]{memoir}
\fi

\newcommand*{\firstdraft}{1 June 2022}%{14 April 2022}
\newcommand*{\firstpublished}{\firstdraft}
\newcommand*{\updated}{\ifarxiv 18 October 2026\else\today\fi}
\newcommand*{\propertitle}{A probability calibrator\\ for machine-learning classifiers}%\\ {\Large}}
\newcommand*{\pdftitle}{\propertitle}
\newcommand*{\headtitle}{Probability calibrator for classifiers}
\newcommand*{\pdfauthor}{K. Dyrland, A. S. Lundervold, P.G.L.  Porta Mana}
\newcommand*{\headauthor}{Dyrland, Lundervold, Porta Mana}
\newcommand*{\reporthead}{\iffalse\else Open Science Framework \href{https://doi.org/10.31219/osf.io/vct9y}{\textsc{doi}:10.31219/osf.io/vct9y}\fi}% Report number

\PassOptionsToPackage{obeyspaces}{url}
\usepackage[T1]{fontenc}
\input{glyphtounicode} \pdfgentounicode=1

\usepackage[utf8]{inputenx}

\usepackage{textcomp}

\usepackage{amsmath}

\usepackage{mathtools}
\usepackage{empheq}% automatically calls amsmath and mathtools
\newcommand*{\widefbox}[1]{\fbox{\hspace{1em}#1\hspace{1em}}}

\usepackage{amssymb}

\usepackage{amsxtra}

\usepackage[main=british]{babel}

\usepackage[autostyle=false,autopunct=false,english=british]{csquotes}
\setquotestyle{american}

\usepackage{amsthm}
\makeatletter
\def\@endtheorem{\endtrivlist}
\makeatother

\usepackage[shortlabels,inline]{enumitem}
\SetEnumitemKey{para}{itemindent=\parindent,leftmargin=0pt,listparindent=\parindent,parsep=0pt,itemsep=\topsep}
\setlist{itemsep=0pt,topsep=\parsep}
\setlist[enumerate,2]{label=(\roman*)}
\setlist[enumerate]{label=(\alph*),leftmargin=1.5\parindent}
\setlist[itemize]{leftmargin=1.5\parindent}
\setlist[description]{leftmargin=1.5\parindent}
\usepackage[babel,theoremfont,largesc,smallerops,nosymbolsc]{newpx}
\makeatletter
\def\re@DeclareMathSymbol#1#2#3#4{%
    \let#1=\undefined
    \DeclareMathSymbol{#1}{#2}{#3}{#4}}
\re@DeclareMathSymbol{\bigoplusop}{\mathop}{largesymbols}{"4C}
\re@DeclareMathSymbol{\bigotimesop}{\mathop}{largesymbols}{"4E}
\re@DeclareMathSymbol{\sumop}{\mathop}{largesymbols}{"50}
\re@DeclareMathSymbol{\prodop}{\mathop}{largesymbols}{"51}
\re@DeclareMathSymbol{\bigcupop}{\mathop}{largesymbols}{"53}
\re@DeclareMathSymbol{\bigcapop}{\mathop}{largesymbols}{"54}
\re@DeclareMathSymbol{\bigwedgeop}{\mathop}{largesymbols}{"56}
\re@DeclareMathSymbol{\bigveeop}{\mathop}{largesymbols}{"57}
\re@DeclareMathSymbol{\bigtimesop}{\mathop}{largesymbolsPXA}{"10}
\makeatother
\DeclareFontFamily{U}{egreek}{\skewchar\font'177}%
\DeclareFontShape{U}{egreek}{m}{n}{<-6>s*[1]eurm5 <6-8>s*[1]eurm7 <8->s*[1]eurm10}{}%
\DeclareFontShape{U}{egreek}{m}{it}{<->s*[1]eurmo10}{}%
\DeclareFontShape{U}{egreek}{b}{n}{<-6>s*[1]eurb5 <6-8>s*[1]eurb7 <8->s*[1]eurb10}{}%
\DeclareFontShape{U}{egreek}{b}{it}{<->s*[1]eurbo10}{}%
\DeclareSymbolFont{egreeki}{U}{egreek}{m}{it}%
\SetSymbolFont{egreeki}{bold}{U}{egreek}{b}{it}% from the amsfonts package
\DeclareSymbolFont{egreekr}{U}{egreek}{m}{n}%
\SetSymbolFont{egreekr}{bold}{U}{egreek}{b}{n}% from the amsfonts package
\DeclareFontFamily{U}{egreekx}{\skewchar\font'177}
\DeclareFontShape{U}{egreekx}{m}{n}{%
       <-7.5>s*[0.9]euex7%
    <7.5-8.5>s*[0.9]euex8%
    <8.5-9.5>s*[0.9]euex9%
    <9.5->s*[0.9]euex10%
}{}
\DeclareSymbolFont{egreekx}{U}{egreekx}{m}{n}
\DeclareMathSymbol{\sumop}{\mathop}{egreekx}{"50}
\DeclareMathSymbol{\prodop}{\mathop}{egreekx}{"51}
\DeclareMathSymbol{\coprodop}{\mathop}{egreekx}{"60}
\makeatletter
\def\sum{\DOTSI\sumop\slimits@}
\def\prod{\DOTSI\prodop\slimits@}
\def\coprod{\DOTSI\coprodop\slimits@}
\makeatother
\DeclareMathSymbol{\varpartial}{\mathalpha}{egreeki}{"40}
\DeclareMathSymbol{\omicron}{\mathalpha}{egreeki}{"6F}
\DeclareMathSymbol{\varAlpha}{\mathalpha}{egreeki}{"41}
\DeclareMathSymbol{\varBeta}{\mathalpha}{egreeki}{"42}
\DeclareMathSymbol{\varEpsilon}{\mathalpha}{egreeki}{"45}
\DeclareMathSymbol{\varZeta}{\mathalpha}{egreeki}{"5A}
\DeclareMathSymbol{\varEta}{\mathalpha}{egreeki}{"48}
\DeclareMathSymbol{\varIota}{\mathalpha}{egreeki}{"49}
\DeclareMathSymbol{\varKappa}{\mathalpha}{egreeki}{"4B}
\DeclareMathSymbol{\varMu}{\mathalpha}{egreeki}{"4D}
\DeclareMathSymbol{\varNu}{\mathalpha}{egreeki}{"4E}
\DeclareMathSymbol{\varOmicron}{\mathalpha}{egreeki}{"4F}
\DeclareMathSymbol{\varRho}{\mathalpha}{egreeki}{"50}
\DeclareMathSymbol{\varTau}{\mathalpha}{egreeki}{"54}
\DeclareMathSymbol{\varChi}{\mathalpha}{egreeki}{"58}
\DeclareMathSymbol{\deltaup}{\mathalpha}{egreekr}{"0E}
\DeclareMathSymbol{\piup}{\mathalpha}{egreekr}{"19}
\usepackage[scaled=0.84]{DejaVuSansMono}% typewriter text
\usepackage[scaled=0.84]{DejaVuSansCondensed}% sans-serif font
\DeclareMathAlphabet{\mathsf}  {T1}{\sfdefault}{m}{sl}
\SetMathAlphabet{\mathsf}{bold}{T1}{\sfdefault}{b}{sl}

\usepackage{upquote}

\usepackage{mathdots}

\usepackage{xcolor}
\definecolor{blue}{HTML}{4477AA}
\definecolor{cyan}{HTML}{66CCEE}
\definecolor{green}{HTML}{228833}
\definecolor{yellow}{HTML}{CCBB44}
\definecolor{red}{HTML}{EE6677}
\definecolor{purple}{HTML}{AA3377}
\definecolor{grey}{HTML}{BBBBBB}
\definecolor{midgrey}{HTML}{888888}
\definecolor{darkgrey}{HTML}{555555}
\definecolor{lgrey}{HTML}{DDDDDD}
\colorlet{shadecolor}{lgrey}

\usepackage{bm}

\usepackage{microtype}

\usepackage[backend=biber,mcite,%subentry,
citestyle=authoryear-comp,bibstyle=pglpm_latex/pglpm-authoryear,dashed=false,%
autopunct=false,sorting=ny,sortcites=false,natbib=false,maxcitenames=2,maxbibnames=8,mincitenames=1,minbibnames=8,giveninits=true,uniquename=false,uniquelist=false,maxalphanames=1,block=space,hyperref=true,defernumbers=false,useprefix=true,sortupper=false,language=british,parentracker=false,autocite=footnote]{biblatex}
\DeclareSortingTemplate{ny}{\sort{\field{sortname}\field{author}\field{editor}}\sort{\field{year}}}
\DeclareFieldFormat{postnote}{#1}
\DefineBibliographyExtras{british}{}
\renewcommand*{\finalnamedelim}{\addspace\amp\space}
\DeclareDelimFormat{multicitedelim}{\addsemicolon\addspace\space}
\DeclareDelimFormat{compcitedelim}{\addsemicolon\addspace\space}
\DeclareDelimFormat{postnotedelim}{\addspace}
\ifarxiv\else\fi
\defbibheading{bibliography}[\bibname]{\section*{#1}\addcontentsline{toc}{section}{#1}%\markboth{#1}{#1}
}
\newcommand*{\citep}{\footcites}
\providecommand{\href}[2]{#2}

\newcommand*{\amp}{\&}
\makeatletter
\newcommand*{\citein}[2][]{\textnormal{\togglefalse{blx@bibliography}\blx@initunit\cite[#1]{#2}}%\addtocategory{extras}{#2}
}
\newcommand*{\citebi}[2][]{\togglefalse{blx@bibliography}\blx@initunit\cite[#1]{#2}%\addtocategory{extras}{#2}
}
\makeatother
\newcommand*{\subtitleproc}[1]{}

\usepackage{xurl}
\PassOptionsToPackage{hyphens}{url}\usepackage[hypertexnames=false,pdfencoding=unicode,psdextra]{hyperref}
\makeatletter
\g@addto@macro{\UrlBreaks}{%
\do\0\do\1\do\2\do\3\do\4\do\5\do\6\do\7\do\8\do\9\do\a\do\b\do\c\do\d\do\e\do\f\do\g\do\h\do\i\do\j\do\k\do\l\do\m\do\n\do\o\do\p\do\q\do\r\do\s\do\t\do\u\do\v\do\w\do\x\do\y\do\z\do\A\do\B\do\C\do\D\do\E\do\F\do\G\do\H\do\I\do\J\do\K\do\L\do\M\do\N\do\O\do\P\do\Q\do\R\do\S\do\T\do\U\do\V\do\W\do\X\do\Y\do\Z%
}
\makeatother

\newcommand*{\doi}[1]{%
\href{https://doi.org/#1}{doi:\allowbreak\nolinkurl{#1}}%
}
\newcommand*{\zenododoi}[1]{%
Zenodo \doi{#1}% 10.31219/osf.io/
}
\newcommand*{\osfdoi}[1]{%
OSF \doi{#1}% 10.31219/osf.io/
}
\newcommand*{\arxivdoi}[1]{%
arXiv \doi{#1}% 10.48550/arXiv.
}
\newcommand*{\biorxivdoi}[1]{%
bioRxiv \doi{#1}% 10.1101/
}
\newcommand*{\halurl}[1]{%
HAL \url{https://#1}% https://hal.science/
}
\newcommand*{\philsciurl}[1]{%
PhilSci \url{#1}% https://philsci-archive.pitt.edu/id/eprint/
}
\newcommand*{\mparcurl}[1]{%
mp\_arc \url{#1}% https://web.ma.utexas.edu/mp_arc-bin/mpa?yn=
}
\newcommand*{\osfproj}[1]{%
OSF \doi{10.17605/osf.io/#1}%
}

\usepackage{graphicx}

\usepackage{pdfrender}

\usepackage[depth=4]{bookmark}
\hypersetup{%
colorlinks=true,
bookmarksnumbered,pdfborder={0 0 0.25},
citebordercolor=blue,citecolor=blue,linkbordercolor=blue,linkcolor=blue,urlbordercolor=blue,urlcolor=blue,breaklinks=true,pdftitle={\pdftitle},pdfauthor={\pdfauthor}}

\usepackage{tensor}

\usepackage[british]{datetime2}
\DTMnewdatestyle{mydate}%
{% definitions
}
\DTMsetdatestyle{mydate}

\ifafour\setstocksize{297mm}{210mm}%{*}% A4
\else\setstocksize{210mm}{5.5in}%{*}% 210x139.7
\fi
\settrimmedsize{\stockheight}{\stockwidth}{*}
\setlxvchars[\normalfont] %313.3632pt for a 66-characters line
\setxlvchars[\normalfont]
\ifafour\settypeblocksize{*}{32pc}{1.618} % A4
\else\settypeblocksize{*}{26.1408pc}{1.618}% nearer to a 66-line newpx and preserves GR
\fi
\setulmargins{*}{*}{1}%gives equal margins
\setlrmargins{*}{*}{*}
\setheadfoot{\onelineskip}{2.5\onelineskip}
\setheaderspaces{*}{2\onelineskip}{*}
\setmarginnotes{2ex}{10mm}{0pt}
\checkandfixthelayout[nearest]
\newenvironment{contributions}{\section*{Author contributions}\addcontentsline{toc}{section}{Author contributions}}{\par}
\newenvironment{acknowledgements}{\section*{Thanks}\addcontentsline{toc}{section}{Thanks}}{\par}
\makeatletter\renewcommand{\appendix}{\par
  \bigskip{\centering
   \interlinepenalty \@M
   \normalfont
   \printchaptertitle{\sffamily\appendixpagename}\par}
  \setcounter{section}{0}%
  \gdef\@chapapp{\appendixname}%
  \gdef\thesection{\@Alph\c@section}%
  \anappendixtrue}\makeatother
\counterwithout{section}{chapter}
\setsecnumformat{\upshape\csname the#1\endcsname\quad}
\setsecheadstyle{\large\bfseries\sffamily%
\centering}
\setsubsecheadstyle{\bfseries\sffamily%
\raggedright}
\setsubsecindent{0pt}%0ex plus 1ex minus .2ex}
\setparaheadstyle{\bfseries\sffamily%
\raggedright}
\copypagestyle{manaart}{plain}
\makeheadrule{manaart}{\headwidth}{0.5\normalrulethickness}
\makeoddhead{manaart}{%
{\footnotesize%\sffamily%
\scshape\headauthor}}{}{{\footnotesize\sffamily%
\headtitle}}
\makeoddfoot{manaart}{}{\thepage}{}
\definecolor{mygray}{gray}{0.333}
\iftypodisclaim%
\ifafour\newcommand\addprintnote{\begin{picture}(0,0)%
\put(245,149){\makebox(0,0){\rotatebox{90}{\tiny\color{mygray}\textsf{This
            document is designed for screen reading and
            two-up printing on A4 or Letter paper}}}}%
\end{picture}}% A4
\else\newcommand\addprintnote{\begin{picture}(0,0)%
\put(176,112){\makebox(0,0){\rotatebox{90}{\tiny\color{mygray}\textsf{This
            document is designed for screen reading and
            two-up printing on A4 or Letter paper}}}}%
\end{picture}}\fi%afourtrue
\makeoddfoot{plain}{}{\makebox[0pt]{\thepage}\addprintnote}{}
\else
\makeoddfoot{plain}{}{\makebox[0pt]{\thepage}}{}
\fi%typodisclaimtrue
\makeoddhead{plain}{\scriptsize\reporthead}{}{}
\pretitle{\begin{center}\LARGE\sffamily%
\bfseries}
\posttitle{\bigskip\end{center}}

\makeatletter\newcommand*{\atf}{\includegraphics[totalheight=\heightof{@}]{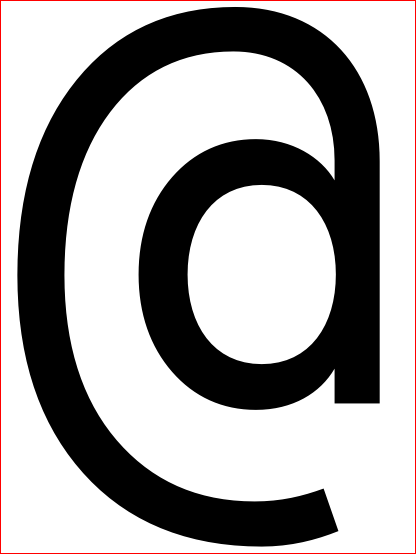}}\makeatother

\providecommand{\epost}[1]{\texttt{\footnotesize\textless#1\textgreater}}
\providecommand{\email}[2]{\href{mailto:#1ZZ@#2 ((remove ZZ))}{#1\protect\atf#2}}

\preauthor{\vspace{-0\baselineskip}\begin{center}
\normalsize\sffamily%
\lineskip  0.5em}
\postauthor{\par\end{center}}
\predate{\DTMsetdatestyle{mydate}\begin{center}\footnotesize}
\postdate{\end{center}\vspace{-2\medskipamount}}

\setfloatadjustment{figure}{\footnotesize}
\captiondelim{\quad}
\captionnamefont{\footnotesize\sffamily%
}
\captiontitlefont{\footnotesize}
\midsloppy
\paragraphfootnotes
\footmarkstyle{\textsuperscript{%\color{red}
\scriptsize\bfseries#1}~}
\colorlet{notecolour}{green}
\title{\propertitle}
\author{%
  K. Dyrland \href{https://orcid.org/0000-0002-7674-5733}{\protect\includegraphics[scale=0.16]{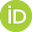}}\\[-\jot]
  {\scriptsize\epost{\email{kjetil.dyrland}{gmail.com}}}%
\\[\jot]%
A. S. Lundervold \href{https://orcid.org/0000-0001-8663-4247}{\protect\includegraphics[scale=0.16]{pglpm_latex/orcid_32x32.png}}\textsuperscript{\ensuremath{\dagger}} \\[-\jot]
{\scriptsize\epost{\email{alexander.selvikvag.lundervold}{hvl.no}}}%
\\[\jot]%
P.G.L.  Porta Mana  \href{https://orcid.org/0000-0002-6070-0784}{\protect\includegraphics[scale=0.16]{pglpm_latex/orcid_32x32.png}}\\[-\jot]
{\scriptsize\epost{\email{pgl}{portamana.org}}}%
\\{\tiny(listed alphabetically)}
\\[\jot]
{\footnotesize Dept of Computer science, Electrical Engineering and Mathematical Sciences,\\Western Norway University of Applied Sciences, Bergen, Norway
  \\[\jot]\textsuperscript{\ensuremath{\dagger}}\amp\ Mohn Medical Imaging and Visualization Centre, Dept of Radiology,\\Haukeland University Hospital, Bergen, Norway\\
}
}

\date{\textbf{Draft}. \firstpublished; updated \updated}

\newcommand*{\pu}{\piup}%constant pi
\newcommand*{\delt}{\deltaup}%Kronecker, Dirac
\newcommand*{\I}{\ensuremath{\mathrm{i}}}
\newcommand*{\e}{\ensuremath{\mathrm{e}}}
\newcommand*{\di}{\mathop{}\!\mathrm{d}}%differential
\newcommand*{\RR}{\bm{\mathrm{R}}}
\newcommand*{\defd}{\coloneqq}

\newcommand*{\dotv}{\mathord{\,\cdot\,}}%variable place
\DeclarePairedDelimiter\clcl{[}{]}
\DeclarePairedDelimiter\set{\{}{\}} %}
\newcommand*{\p}{\mathrm{p}}%probability
\renewcommand*{\P}{\mathrm{P}}%probability
\renewcommand*{\|}[1][]{\nonscript\:#1\vert\nonscript\:\mathopen{}}
\newcommand*{\mo}[1][=]{\mathclose{}\mathord{\nonscript\mkern0.5mu#1\nonscript\mkern0.5mu}\mathopen{}}
\newcommand*{\sect}{\S}% Sect.~
\newcommand*{\sects}{\S\S}% Sect.~
\newcommand*{\chap}{ch.}%
\newcommand*{\chaps}{chs}%
\newcommand*{\eqn}{eq.}%
\newcommand*{\fig}{fig.}%
\newcommand*{\figs}{figs}%

\newcommand*{\eg}{{e.g.}}

\newcommand*{\ie}{{i.e.}}
\newcommand*{\cf}{{cf.}}
\DeclareMathOperator*{\argmax}{arg\,max}

\newcommand*{\eu}{\bar{U}}

\newcommand*{\wf}{w}
\newcommand*{\wfo}{w_{\textrm{g}}}
\newcommand*{\tI}{\textit{I}}
\newcommand*{\tII}{\textit{II}}

\newcommand*{\texts}[1]{\text{\small #1}}
\newcommand*{\ml}{machine-learning}
\newcommand*{\RF}{random forest}
\newcommand*{\rf}{random-forest}
\newcommand*{\CNN}{convolutional neural network}
\newcommand*{\cnn}{convolutional-neural-network}
\newcommand*{\br}{r}
\newcommand*{\No}{\mathrm{N}}
\newcommand*{\umatrix}[4]{\begin{bmatrix*}[r]#1&#2\\#3&#4\end{bmatrix*}}
\newcommand*{\sumatrix}[4]{\begin{bsmallmatrix*}[r]#1&#2\\#3&#4\end{bsmallmatrix*}}
\begin{document}
\captiondelim{\quad}\captionnamefont{\footnotesize}\captiontitlefont{\footnotesize}
\selectlanguage{british}\frenchspacing
\maketitle

%%%%%%%%%%%%%%%%%%%%%%%%%%%%%%%%%%%%%%%%%%%%%%%%%%%%%%%%%%%%%%%%%%%%%%%%%%%%
%%% Abstract
%%%%%%%%%%%%%%%%%%%%%%%%%%%%%%%%%%%%%%%%%%%%%%%%%%%%%%%%%%%%%%%%%%%%%%%%%%%%
\abstractrunin
\abslabeldelim{}
\renewcommand*{\abstractname}{}
\renewcommand*{\abstracttextfont}{\normalfont\footnotesize}
\setlength{\abstitleskip}{-\absparindent}
\begin{abstract}\labelsep 0pt%
  \noindent
  In fields such as medicine and drug discovery, the ultimate goal of a classification is not to guess a class, but to choose the optimal course of action among a set of possible ones, usually not in one-one correspondence with the set of classes. This decision-theoretic problem requires proper probabilities for the classes. Proper probabilities conditional on the features are computationally almost impossible to find in many important cases. The main idea of the present work is to calculate probabilities conditional not on the features, but on the trained classifier's output. This calculation is cheap, needs to be made only once, and provides an output-to-probability \enquote{calibrator} or \enquote{transducer} that can be applied to all future outputs of the classifier. In conjunction with problem-dependent utilities, the probabilities of the transducer allow us to find the optimal choice among the classes or among a set of more general decisions, by means of expected-utility maximization. This idea is demonstrated in a simplified drug-discovery problem with a highly imbalanced dataset. The transducer and utility maximization together always lead to improved results, sometimes close to the theoretical maximum, for all sets of problem-dependent utilities. The one-time-only calculation of the transducer also provides, automatically: (i) a quantification of the uncertainty about the transducer itself; (ii) the expected utility of the augmented algorithm (including its uncertainty), which can be used for algorithm selection; (iii) the possibility of using the algorithm in a \enquote{generative mode}, useful if the training dataset is biased.
% It is argued that the optimality, flexibility, and uncertainty assessment provided by the transducer \amp\ augmentation are dearly needed for classification problems in fields such as medicine and drug discovery.
  % The present work sells probabilities for \ml\ algorithms at a bargain price. It shows that these probabilities are just what those algorithms need in order to increase their sales.
% \\\noindent\emph{\footnotesize Note: Dear Reader
%     \amp\ Peer, this manuscript is being peer-reviewed by you. Thank you.}
% \par%\\[\jot]
% \noindent
% {\footnotesize PACS: ***}\qquad%
% {\footnotesize MSC: ***}%
%\qquad{\footnotesize Keywords: ***}
\end{abstract}
\selectlanguage{british}\frenchspacing

%%%%%%%%%%%%%%%%%%%%%%%%%%%%%%%%%%%%%%%%%%%%%%%%%%%%%%%%%%%%%%%%%%%%%%%%%%%%
%%% Epigraph
%%%%%%%%%%%%%%%%%%%%%%%%%%%%%%%%%%%%%%%%%%%%%%%%%%%%%%%%%%%%%%%%%%%%%%%%%%%%
% \asudedication{\small ***}
% \vspace{\bigskipamount}
% \setlength{\epigraphwidth}{.7\columnwidth}
% %\epigraphposition{flushright}
% \epigraphtextposition{flushright}
% %\epigraphsourceposition{flushright}
% \epigraphfontsize{\footnotesize}
% \setlength{\epigraphrule}{0pt}
% %\setlength{\beforeepigraphskip}{0pt}
% %\setlength{\afterepigraphskip}{0pt}
% \epigraph{\emph{text}}{source}

%%%%%%%%%%%%%%%%%%%%%%%%%%%%%%%%%%%%%%%%%%%%%%%%%%%%%%%%%%%%%%%%%%%%%%%%%%%%
%%% BEGINNING OF MAIN TEXT
%%%%%%%%%%%%%%%%%%%%%%%%%%%%%%%%%%%%%%%%%%%%%%%%%%%%%%%%%%%%%%%%%%%%%%%%%%%%

\section{The inadequacy of common classification approaches}
\label{sec:goal_class}

As the potential of using \ml\ algorithms in important fields such as medicine or drug discovery increases \autocites{lundervoldetal2019,chenetal2018,green2019}, the \ml\ community ought to keep in mind what the actual needs and inference contexts in such fields are. We must avoid trying (intentionally or unintentionally) to convince such fields to change their needs, or to ignore their own contexts just to fit \ml\ solutions that are available and fashionable at the moment. Rather, we must make sure that the solutions fit needs \amp\ context, and amend them if they do not.

The \ml\ mindset and approach to problems such as classification in such new important fields is often still inadequate in many respects. It reflects simpler needs and contexts of many  inference problems successfully tackled by machine learning earlier on.

A stereotypical \enquote{cat vs dog} image classification, for instance, has four very important differences from a \enquote{disease \tI\ vs disease \tII} medical classification, or from an \enquote{active vs inactive} drug classification:
\begin{enumerate}[label=(\roman*)]

\item\label{item:gain_loss}Nobody presumably dies or loses large amounts of money if a cat image is misclassified as dog or vice versa. But a person can die if a disease is misdiagnosed; huge capitals can be lost if an ultimately ineffective drug candidate is pursued. The \emph{gains and losses} -- or generally speaking the \emph{utilities} -- of correct and incorrect classifications in the former problem and in the two latter problems are vastly different.

\item\label{item:decisions_classes} To what purpose do we try to guess whether an image's subject is a cat or a dog? For example because we must decide whether to put it in the folder \enquote{cats} or in the folder \enquote{dogs}. To what purpose do we try to guess a patient's disease or a compound's chemical activity? A clinician does not simply tell a patient \enquote*{You probably have such-and-such disease. Goodbye!}, but has to decide among many different kinds of treatments. The candidate drug compound may be discarded, pursued as it is, modified, and so on. \emph{The ultimate goal of a classification is always some kind of decision}, not just a class guess. In the cat-vs-dog problem there is a natural one-one correspondence between classes and decisions. But in the medical or drug-discovery problems \emph{the set of classes and the set of decisions are very different}, and have even different numbers of elements.

\item\label{item:optimal_truth} If there is a 70\% probability that an image's subject is a cat, then it is natural to put it in the folder \enquote{cats} rather than \enquote{dogs} (if the decision is only between these two folders). If there is a 70\% probability that a patient has a particular health condition, it may nonetheless be better to dismiss the patient -- that is, to behave as if there was no condition. This is the optimal decision, for example, when the only available treatment for the condition would severely harm the patient if the condition were not present. Such treatment would be recommended only if the probability for the condition were much higher than 70\%. Similarly, even if there is a 70\% probability that a candidate drug is active it may nonetheless be best to discard it. This is the economically most advantageous choice if pursuing a false-positive leads to large economic losses. \emph{The target of a classification is not what’s most probable, but what’s optimal.}
  % \mynotep{add something on proper probabilities, connecting to~\ref{item:determ_statist}}

\item\label{item:determ_statist} The relation from image pixels to house-pet subject may be almost deterministic; so we are effectively looking for or extrapolating a function $\texts{pet} \mo f(\texts{pixels})$ contaminated by little noise. But the relation between medical-test scores or biochemical features on one side, and disease or drug activity on the other, is typically \emph{probabilistic}; so a function $\texts{disease} \mo f(\texts{scores})$
  % scores${}\!\mapsto\!{}$disease
  or $\texts{activity} \mo f(\texts{features})$
  % features${}\!\mapsto\!{}$activity
  does not even exist. \emph{We are assessing statistical relationships} $\P(\texts{disease} , \texts{scores})$ or $\P(\texts{activity} , \texts{features})$ instead, which include deterministic ones as special cases.

\end{enumerate}

In summary, there is place to improve classifiers so as to
\begin{enumerate*}[label=(\roman*)]
\item quantitatively take into account actual utilities,
\item separate classes from decisions,
\item target optimality rather than \enquote{truth}, %-- which cannot be targeted because it is unknown.
\item output and use proper probabilities.
\end{enumerate*}
% \begin{enumerate*}[label=(\roman*)]
% \item neglect actual gains and losses or treat them only qualitatively,
% \item neglect proper probabilities and often demotes density regression (statistical inference) to functional regression (interpolation),
% \item confuse classes and decisions,
% \item mistake truth for optimality.
% \end{enumerate*}

\medskip

In artificial intelligence and machine learning it is known how to address all these issues in principle -- the theoretical framework is for example beautifully presented in the first 18 chapters or so of Russell \amp\ Norvig's \cites*{russelletal1995_r2022} text \autocites[see also][]{selfetal1987,cheeseman1988,cheeseman1995_r2018,pearl1988,mackay1995_r2005}.

Issues~\ref{item:gain_loss}--\ref{item:optimal_truth} are simply solved by adopting the standpoint of \emph{Decision Theory}, which we briefly review in \sect~\ref{sec:utility_classification} below and discuss at length in a companion work \autocites{dyrlandetal2022}. In short: The set of decisions and the set of classes pertinent to a problem are separated if necessary. A utility is associated to each decision, relative to the occurrence of each particular class; these utilities are assembled into a utility matrix: one row per decision, one column per class. This matrix is multiplied by a column vector consisting in the probability for the classes. The resulting vector of numbers contains the expected utility of each decision. Finally we select the decision having \emph{maximal expected utility}, according to the principle bearing this name. Such procedure also takes care of the class imbalance problem \autocites[\cf\ the analysis by][(they use the term \enquote{cost} instead of \enquote{utility})]{drummondetal2005}.

Clearly this procedure is computationally inexpensive and ridiculously easy to implement in any \ml\ classifier. The difficulty is that this procedure requires \emph{sensible probabilities} \autocites[\enquote*{credibilities \textins*{that} would be agreed by all rational men if there were any rational men}][]{good1966} for the classes, which brings us to issue~\ref{item:determ_statist}, the most difficult.

Some machine-learning algorithms for classification, such as support-vector machines, output only a class label. Others, such as deep networks, output a set of real numbers that can bear some qualitative relation to the plausibilities of the classes. But these numbers cannot be reliably interpreted as proper probabilities, that is, as the degrees of belief assigned to the classes by a rational agent \autocites{mackay1992d,galetal2016}[\chaps~2, 12, 13]{russelletal1995_r2022}; or, in terms of \enquote{populations} \autocites{lindleyetal1981}%[\sect~II.4]{fisher1956_r1967}
, as the expected frequencies of the classes in the hypothetical population of units (degrees of belief and frequencies being related by de~Finetti's theorem \autocites[\chap~4]{bernardoetal1994_r2000}{dawid2013}). Algorithms that internally do perform probabilistic calculations, for instance naive-Bayes or logistic-regression classifiers \autocites[\sect~3.5, \chap~8]{murphy2012}[\sects~8.2, 4.3]{bishop2006}[\chap~10, \sect~17.4]{barber2007_r2020}, unfortunately rest on strong probabilistic assumptions, such as independence and particular shapes of distributions, that are often unrealistic (and their consistency with the specific application is rarely checked). Only particular classifiers such as Bayesian neural networks \autocites{nealetal2006}[\sect~5.7]{bishop2006} output sensible probabilities, but they are computationally very expensive. The stumbling block is the extremely high dimensionality of the feature space, which makes the calculation of the conditional probabilities
\[
  \P(\texts{class}, \texts{feature} \| \texts{training data})
\]
(a problem opaquely called \enquote{density regression} or \enquote{density estimation}\autocites{ferguson1983,thorburn1986,hjort1996,dunsonetal2007}) computationally unfeasible.

If we solved the issue of outputting proper probabilities then the remaining three issues would be easy to solve, as discussed above.

\bigskip

In the present work we propose an alternative solution to calculate proper class probabilities, which can then be used in conjunction with utilities to perform the final classification or decision.

This solution consists in a sort of \enquote{calibrator} or \enquote{transducer} that transforms the algorithm's raw output into a probability. This probability transducer has a low computational cost, can be applied to all commonly used classifiers and to simple regression algorithms, does not need any changes in algorithm architecture or in training procedures, and is grounded on first principles. The probability obtained from the transducer can be combined with utilities to perform the final classification or decision task. Notably, the set of available decisions and their utilities \emph{can be changed on the fly}, for each new classification instance.

This probability transducer has also three other great benefits, which come automatically with its computation. First, it gives a quantification of how much the probability would change if we had further data to calculate the transducer. Second, it can give an \emph{evaluation of the whole classifier} -- including an uncertainty about such evaluation -- that allows us to compare it with other classifiers and choose the optimal one. Third, it allows us to calculate both the probability of class conditional on features, and \emph{the probability of features conditional on class}. In other words it allows us to use the classification algorithm in both \enquote{discriminative} and \enquote{generative} modes \autocites[\sect~21.2.3]{russelletal1995_r2022}[\sect~8.6]{murphy2012}, even if the algorithm was not designed for a generative use.

In \sect~\ref{sec:transducer} we present the general idea behind the probability transducer and its calculation. Its combination with the rule of expected-utility maximization to perform classification is discussed in \sect~\ref{sec:utility_classification}; we call this combined use the \enquote{augmentation} of a classifier.

In \sect~\ref{sec:demonstration} we demonstrate the implementation, use, and benefits of classifier augmentation in a concrete drug-discovery classification problem and dataset, with a \RF\ and a \CNN\ classifiers.

Section~\ref{sec:overview_other_uses} offers a synopsis of further benefits and uses of the probability transducer, which are obtained almost automatically from its calculation.

Finally, we give a summary and discussion in \sect~\ref{sec:summary_discussion}, including some teasers of further applications to be discussed in future work.

\section{An output-to-probability transducer}
\label{sec:transducer}

% In the present section we explain the main idea in informal and intuitive terms, and give its  mathematical essentials only. % Stricter logical derivation and more mathematical details are left to appendix\mynotep{\ldots}.

\subsection{Main idea: algorithm output as a proxy for the features}
\label{sec:essential_idea}

Let us first consider the essentials behind a classification (or regression) problem. We have the following quantities:
\begin{itemize}
\item the \emph{feature} values of a set of known units,
\item the \emph{classes} of the same set of units,
\end{itemize}
which together form our \emph{learning} or \emph{training data}; and
\begin{itemize}[resume]
\item the feature value of a \emph{new} unit,
\end{itemize}
where the \enquote{units} could be widgets, images, patients, drug compounds, and so on, depending on the classification problem. From these quantities we would like to infer
\begin{itemize}[resume]
\item  the class of the new unit.
\end{itemize}
This inference consists in probabilities
\begin{equation}
  \label{eq:prob_essential}
  \P(\texts{class of new unit} \| \texts{feature of new unit},\,
  \texts{classes \amp\ features of known units})
\end{equation}
for each possible class.

These probabilities are obtained through the rules of the probability calculus \autocites{jaynes1994_r2003}[\chaps~12--13]{russelletal1995_r2022}[\addcolon see further references in appendix~\ref{sec:maths_transducer}]{gregory2005,hailperin2011,jeffreys1939_r1983}; in this case specifically through the so-called de~Finetti theorem \autocites[\chap~4]{bernardoetal1994_r2000}{dawid2013} which connects training data and new unit. This theorem is briefly summarized in appendix~\ref{sec:deFinetti}.

Combined with a set of utilities, these probabilities allow us to determine an optimal, further decision to be made among a set of alternatives. Note that the inference~\eqref{eq:prob_essential} includes deterministic interpolation, \ie\ the assessment of a function $\texts{class} \mo f(\texts{feature})$, as a special case, when the probabilities are essentially $0$s and $1$s.

A trained classifier should ideally output the probabilities above when applied to the new unit. Machine-learning classifiers trade this capability for computational speed -- with an increase in the latter of several orders of magnitude \autocites[to understand this trade-off in the case of neural-network classifiers see \eg][]{mackay1992,mackay1992b,mackay1992d}[\sect~16.5 esp.~16.5.7]{murphy2012}[see also the discussion by][]{selfetal1987}. Thus their output cannot be considered a probability, but \emph{it still carries information about both class and feature variables}.

% We can introduce the known output variable provided by the classifier in the probability~\eqref{eq:prob_essential}:
% \begin{equation}
%   \label{eq:prob_essential_output}
%   \P(\texts{class of unit} \| \texts{output for unit}, 
%   \texts{feature of unit}, \texts{training data}) \ .
% \end{equation}
% This probability must be numerically equal to~\eqref{eq:prob_essential} because the classifier's output cannot give more information about the class than is already present in the feature and in the training data (the classifier would be biased otherwise).

Our first step is to acknowledge that the information contained in the feature and in the training data, relevant to the class of the new unit, is simply inaccessible to us because of computational limitations. We do have access to the output for the new unit, however, which does carry relevant information. Thus what we can do is to calculate the probability
\begin{empheq}[box=\widefbox]{equation}
  \label{eq:accessible_prob_newunit}
  \P(\texts{class of new unit} \| \texts{output for new unit})
\end{empheq}
for each class.

Once we calculate the numerical values of these conditional probabilities, we effectively have a function that maps the algorithm's output to class probabilities. It therefore acts as an \emph{output-to-probability transducer}.

This idea can also be informally understood in two ways. First: the classifier's output is regarded as a proxy for the feature. Second: the classifier is regarded as something analogous to a \emph{diagnostic test}, such as any common diagnostic or prognostic test used in medicine for example. A diagnostic test is useful because its result has a probabilistic relationship with the unknown of interest, say, a medical condition. This relationship is easier to quantify than the one between the condition and the more complex biological variables that the test is exploiting \enquote{under the hood}. Likewise, the output of a classifier has a probabilistic relationship with the unknown class (owing to the training process); and this relationship is in many cases easier to quantify than the one between the class and the typically complex \enquote{features} that are the classifier's input. We do not take diagnostic-test results at face value -- if a flu test is \enquote{positive} we do not conclude that the patient has the flu -- but rather arrive at a probability that the patient has the flu, given some statistics about results of tests performed on \emph{verified samples} of true-positive and true-negative patients \autocites[\chap~5]{soxetal1988_r2013}[\chap~5]{huninketal2001_r2014}[see also][]{jennyetal2018}. Analogously, we need some calibration data to find the probabilities~\eqref{eq:accessible_prob_newunit}.

\subsection{Calibration data}
\label{sec:calibration_data}

To calculate the conditional probabilities~\eqref{eq:accessible_prob_newunit} it is necessary to have examples of further pairs $(\texts{class of unit}, \texts{output for unit})$, of which the new unit's pair can be considered a \enquote{representative sample} \autocites[for a critical analysis of the sometimes hollow term \enquote{representative sample} see][]{kruskaletal1979,kruskaletal1980} and vice versa -- exactly for the same reason why we need training data in the first place to calculate the probability of a class given the feature. Or, with a more precise term, the examples and the new unit must be \emph{exchangeable} \autocites{lindleyetal1981}.

For this purpose, can we use the pairs $(\texts{class of unit}, \texts{output for unit})$ of the training data? This would be very convenient, as those pairs are readily available. But answer is no. The reason is that the outputs of the training data are produced from the features \emph{and the classes} jointly; this is the very point of the training phase. There is therefore a direct informational dependence between the classes and the outputs of the training data. For the new unit, on the other hand, the classifier produces its output from the feature alone. \emph{As regards the probabilistic relation between class and output, the new unit is not exchangeable with (or a representative sample of) the training data}.

We need a data set where the outputs are generated by simple application of the algorithm to the feature, as it would occur in its concrete use, and the classes are known. The \emph{test data} of standard \ml\ procedures are exactly what we need. The new unit can be considered exchangeable with the test data. We rename such data \enquote{transducer-calibration data}, owing to its new purpose.

The probability we want to calculate is therefore
\begin{equation}
  \label{eq:prob_output}
  \P(\texts{class of new unit} \| \texts{output for new unit},\,
  \texts{classes \amp\ outputs of calibr. data}) \ .
\end{equation}
For classification algorithms that output a quantity much simpler than the features, like a vector of few real components for instance, the probability above can be exactly calculated. Thus, once we obtain the classifier's output for the new unit, we can calculate a probability for the new unit's class.

The probability values~\eqref{eq:prob_output}, for a fixed class and variable output, constitute a sort of \enquote{calibration curve} (or hypersurface for multidimensional outputs) of the output-to-probability transducer for the classifier. See the concrete examples of \figs~\ref{fig:prob_curve_RF} on page~\pageref{fig:prob_curve_RF}, and~\ref{fig:prob_curve_CNN} on page~\pageref{fig:prob_curve_CNN}. It must be stressed that such curve needs to be calculated only once, and it can be used for all further applications of the classifier to new units.

\medskip

What is the relation between the ideal incomputable probability~\eqref{eq:prob_essential} and the probability~\eqref{eq:prob_output} obtained by proxy? If the output $y$ of the classifier is already very close to the ideal probability~\eqref{eq:prob_essential}, or a monotonic function thereof, isn't the proxy probability~\eqref{eq:prob_output} throwing it away and replacing it with something different? Quite the opposite. Owing to de~Finetti's theorem, if the output $y$ is almost identical with the ideal probability, then it becomes increasingly close to the frequency distribution of the training data, as their number increases (see appendix~\ref{sec:deFinetti}); the same happens with the proxy probability and the frequency distribution of the calibration data. But these two data sets should be representative of each other and of future data -- otherwise we would be \enquote{learning} from irrelevant data -- and therefore their frequency distributions should also converge to each other. Consequently, by transitivity we expect the proxy probability to become increasingly close to the output $y$. Actually, if the output is not exactly the ideal probability~\eqref{eq:prob_essential} but a monotonic function of it, the proxy probability~\eqref{eq:prob_output} will reverse such monotonic relationship, giving us back the ideal probability.

Obviously all these considerations only hold if we have good training and calibration sets, exchangeable with (representative of) the real data that will occur in our application.

% It can be shown that the two are related by convex combination or mixing:
% \begin{multline}
%   \label{eq:relation_two_probs}
%   \P(\texts{class} \| \texts{output},\,
%   \texts{classes \amp\ outputs for calibration data}) ={}\\[\jot]
%   \sum_{\mathclap{\substack{\text{feature}\\\text{training data}}}}
%   w(\texts{feature}, \texts{training data})\times
%   \P(\texts{class} \| \texts{feature},\, \texts{training data})
% \end{multline}
% where $w$ are positive and normalized functions. This means that the proxy probability is generally less sharp, that is, farther away from $0$ and $1$, than the ideal one. This is an obvious consequence of the loss of information about the feature value. Such mixing, on the other hand, only leads to more conservative probabilities, not to over-confident ones.
% \mynotep{add note about the fact that the goodness of the results still greatly depends on the goodness of the classifier.}

\medskip

Since we are using as calibration data the data traditionally set aside as \enquote{test data} instead, an important question arises. Do we then need a third, separate test dataset for the final evaluation and comparison of candidate classifiers or hyperparameters? This would be inconvenient: it would reduce the amount of data available for training.

The answer is no: \emph{the calibration set automatically also acts as a test set}. In fact, from the calculations for the probability transducer, discussed in the next section, we can also arrive at a final evaluation value for the algorithm as a whole. See \sect~\ref{sec:yield_of_classifier} and appendix~\ref{sec:algorithm_yield} for more details about this.

It may be useful to explain why this is the case, especially for those who may mistakenly take for granted the universal necessity of a test set. Many standard \ml\ methodologies need a test set because they are only an approximation of the ideal inference performed with the probability calculus. The latter needs no division of available data into different sets: something analogous to such division is automatically made internally, so to speak (see appendix~\ref{sec:deFinetti}). It can be shown \autocites{portamana2019b,fongetal2020,wald1949}[many examples of this fact are scattered across the text by][]{jaynes1994_r2003} that the mathematical operations behind the probability rules correspond to making \emph{all possible} divisions of available data between \enquote{training} and \enquote{test}, as well as \emph{all possible} cross-validations with folds of all orders. It is this completeness and thoroughness that makes the ideal inference by means of the probability calculus almost computationally impossible in some cases. We thus resort to approximate but faster \ml\ methods. These methods do not typically perform such data partitions \enquote{internally} and automatically, so we need to make them -- and only approximately -- by hand.

\medskip

Let us stress that the performance of a classifier equipped with a probability transducer still depends on the training of the raw classifier, which is the stage where a probabilistic relation between output and class is established. If the classifier's output has no mutual information with the true class (their probabilities are essentially independent), then the transducer will simply yield a uniform probability over the classes.

The question then arises of what is the optimal division of available data into the training set and the calibration set. If the calibration set is too small, the transducer curve is unreliable. If the training set is too small, the correlation between output and class is unreliable. In future work we would like to find the optimal balance, possibly by a first-principle calculation.

\subsection{Calculation of the probabilities}
\label{sec:calculation_transducer}

Let us denote by $c$ the class value of a new unit, by $y$ the output of the classifier for the new unit, and by $D \defd \set{c_{i}, y_{i}}$ the classes and classifier outputs for the transducer-calibration data.

It is more convenient to focus on the \emph{joint} probability of class and output given the data,
\begin{equation}
  \label{eq:prob_output}
  \p(c, y \| D) \ ,
\end{equation}
rather than on the conditional probability of the class given the output and calibration data,~\eqref{eq:prob_output}.

The joint probability is calculated using standard non-parametric Bayesian methods \autocites[for introductions and reviews see \eg][]{walker2013,muelleretal2004b,hjort1996}. \enquote{Non-parametric} in this case means that we do not make any assumptions about the shape of the probability curve as a function of $c,y$ (contrast this with logistic regression, for instance), or about special independence between the variables (contrast this with naive-Bayes). The only assumption made -- and we believe it is quite realistic -- is that the curve must have some minimal degree of smoothness. This assumption allows for much leeway, however: \figs~\ref{fig:prob_curve_RF} and \ref{fig:prob_curve_CNN_section} for instance show that the probability curve can still have very sharp bends, as long as they are not cusps.

Non-parametric methods differ from one another in the kind of \enquote{coordinate system} they select on the infinite-dimensional space of all possible probability curves, that is, in the way they represent a general positive normalized function.

We choose the representation discussed by Dunson \amp\ Bhattacharya\autocites{dunsonetal2011}[see also the special case discussed by][]{rasmussen1999}. The end result of interest in the present section is that the probability density $p(c,y \| D)$, with $c$ discrete and $y$ continuous and possibly multi-dimensional, is expressed as a sum
\begin{equation}
  \label{eq:representation_P}
  p(c, y \| D) = \sum_{k} q_{k}\ A(c \| \alpha_{k})\ B(y \| \beta_{k})
\end{equation}
of a finite but large number of terms \autocites[see][on why the number of terms does not need to be infinite]{ishwaranetal2002b}. Each term is the product of a positive weight $q_{k}$, a probability distribution $A(c \| \alpha_{k})$ for $c$ depending on parameters $\alpha_{k}$, and a probability density $B(c \| \beta_{k})$ for $y$ depending on parameters $\beta_{k}$. These distributions are chosen by us according to convenience; see the appendix~\ref{sec:deFinetti} for further details. The parameter values can be different from term to term, as indicated by the index $k$. The weights $\set{q_{k}}$ are normalized. For simplicity we shall from now on omit the dependence \enquote{$\|\dotsc, D)$} on the calibration data, leaving it implicitly understood.

This mathematical representation can approximate (under some norm) any smooth probability density in $c$ and $y$. It has the advantages of being automatically positive and normalized, and of readily producing the marginal distributions for $c$ and for $y$:
\begin{equation}
  \label{eq:marginals}
  p(c) = \sum_{k} q_{k}\ A(c \| \alpha_{k})\ ,
  \qquad
  p(y) = \sum_{k} q_{k}\ B(y \| \beta_{k}) \ ,
\end{equation}
from which also the conditional distributions are easily obtained:
\begin{subequations} \label{eq:conditionals}
  \begin{align}
    \label{eq:conditional_c_given_y}
  \Aboxed{  p(c \| y) &= \sum_{k} \frac{q_{k}\ B(y \| \beta_{k})
                }{\sum_{l} q_{l}\ B(y \| \beta_{l})}\
                A(c \| \alpha_{k})}
    \\
    \label{eq:conditional_y_given_c}
    p(y \| c) &= \sum_{k} \frac{q_{k}\ A(c \| \alpha_{k})
                }{\sum_{l} q_{l}\ A(c \| \alpha_{l})}\
                B(y \| \beta_{k}) \ .
  \end{align}
\end{subequations}

In the rest of the paper we shall use formula~\eqref{eq:conditional_c_given_y}, the probability of the class given the algorithm's output, as typically done with discriminative algorithms. 

\medskip

The weights and parameters $\set{q_{k}, \alpha_{k}, \beta_{k}}$ are the heart of this representation, because the shape of the probability curve $p(c \| y, D)$ depends on their values. They are determined by the test data $D$. Their calculation is done via Markov-chain Monte Carlo sampling, discussed in appendix~\ref{sec:MCMC}. For low-dimensional $y$ and discrete $c$ (or even continuous, low-dimensional $c$, which means we are working with a regression algorithm), this calculation can be done in a matter of hours, and \emph{it only needs to be done once}.

Once calculated, these parameters are saved in memory and can be used to compute any of the probabilities~\eqref{eq:representation_P}, \eqref{eq:marginals}, \eqref{eq:conditionals} as needed, as discussed in the next subsection. Such computations take less than a second.

Note that the role of the classifier in this calculation is simply to produce the outputs $y$ for the calibration data, after having been trained in any standard way on a training data set. No changes in its architecture or in its training procedure have been made, nor are any required.

\section{Utility-based classification}
\label{sec:utility_classification}

We refer to our companion work \cite{dyrlandetal2022}, \sect~2, for a more detailed presentation of decision theory and for references. In the following we assume familiarity with the material presented there.

Our classification or decision problem has a set of decisions, which we can index by $i = 1,2,\dotsc$. As discussed in \sect~\ref{sec:goal_class}, these need not be the same as the possible classes; the two sets may even be different in number. But the true class, which is unknown, determines the \emph{utility} that a decision yields. If we choose decision $i$ and the class $c$ is true, our eventual utility will be $U_{ic}$.\footnote{We apologize for the difference in notation from our companion work, where the class variable is \enquote{$j$} and the utilities \enquote{$U_{ij}$}} These utilities are assembled into a rectangular matrix $(U_{ic})$ with one row per decision and one column per class. Note that the case where decisions and classes are in a natural one-one correspondence, as in the cat-vs-dog classification example of \sect~\ref{sec:goal_class}, is just a particular case of this more general point of view. In such a specific case we may replace \enquote{decision} with \enquote{class} in the following discussion, and the utility matrix is square.

\medskip

Now let us consider the application of the algorithm, with the probabilities calculated in the preceding section, to a new unit.

\begin{enumerate}[label=\arabic*.]
\item Fed the unit's features to the classifier, which outputs the real value $y$.

\item\label{item:calculate_probs} Calculate $p(c \| y)$, for each value of $c$, from formula~\eqref{eq:conditional_c_given_y}, using the parameters $\set{q_{k}, \alpha_{k}, \beta_{k}}$ stored in memory. These are the probabilities of the classes, which are collected in a column vector $(p_{c})$.

\item\label{item:combine_prob_utilities} The expected utility $\eu_{i}$ of decision $i$ is given by the matrix product of an appropriate utility matrix $(U_{ic})$ and the column vector $(p_{c})$:
  \begin{equation}
    \label{eq:expe_utilities}
    \eu_{i} \defd \sum_{c} U_{ic}\ p_{c} \ .
  \end{equation}

\item\label{item:maximize_utility} Choose the decision $i^{*}$ having largest $\eu_{i}$, according to the principle of maximum expected utility:
  \begin{empheq}[box=\widefbox]{equation}
    \label{eq:max_expe_utility}
    \text{choose}\quad
    i^{*} = \argmax_{i}\set*{\eu_{i}} \equiv \argmax_{i}\set[\bigg]{\sum_{c} U_{ic}\ p_{c}}
  \end{empheq}
\end{enumerate}

\medskip

We call the procedure above, especially steps~\ref{item:calculate_probs}--\ref{item:maximize_utility}, the \emph{augmentation} of the classifier.

In step~\ref{item:calculate_probs} we have effectively translated the classifier's raw output into a more sensible probability. From this point of view the function $p(c \| y)$ can be considered as a more appropriate substitute of the softmax function, for instance, at the output of a neural network (compare \fig~\ref{fig:prob_curve_CNN}).

The matrix multiplication of and subsequent selection of steps~\ref{item:combine_prob_utilities}--\ref{item:maximize_utility} are computationally inexpensive; they can be considered as substitutes of the \enquote{argmax} selection that typically happen at the continuous output of a classifier.

It should be noted that the utilities $U_{ic}$ used in step~\ref{item:combine_prob_utilities} can either be the same for each new unit, or different from unit to unit. The augmentation procedure is therefore extremely flexible, at no additional computational cost.

\section{Demonstration}
\label{sec:demonstration}

\subsection{Overview}
\label{sec:demo_overview}

We illustrate the implementation of the probability transducer and its combination with utility-based decisions in a concrete example. The evaluation of the results is also made from the standpoint of decision theory, using utility-based metrics, as explained in our companion paper \autocites{dyrlandetal2022}.

A couple of remarks may clarify the purpose of this illustration and our choice of classification problem.

The internal consistency of decision theory guarantees that utility-based decisions always improve on, or at least give as good results as, any other procedure, including the standard classification procedures used in machine learning. This is intuitively obvious: we are, after all, grounding our single class choices upon the same gains \amp\ losses that underlie our classification problem and that are used in its evaluation. The present illustration is therefore not a proof for such improvement -- none is needed. It is a reassuring safety check, though: if the results were negative it would mean that errors were made in applying the method or in the computations.

Rather than looking for some classification problem and dataset on which the decision-theoretic approach could lead to astounding improvements, we choose one where \ml\ classifiers already give excellent results, therefore difficult to improve upon; and which is characterized by a naturally high class imbalance. The classification problem is moreover of interest to us for other ongoing research projects.

The binary-classification task is a simplified version of an early-stage drug-discovery problem: to determine whether a molecule is chemically \enquote{inactive} (class~$0$) or \enquote{active} (class~$1$) towards one specific target protein.

Two \ml\ classifiers are considered: a Random Forest and a residual Convolutional Neural Network (ResNet), details of which are given in appendix~\ref{sec:appendix_algorithms}. The \RF\ takes as input a set of particular physico-chemical characteristics of a molecule, and outputs a real number in the range $\clcl{0,1}$, corresponding to the fraction of decision trees which vote for class~$1$, \enquote{active}. The \cnn\ takes as input an image representing the chemical and spatial structure of the molecule, and outputs two real numbers roughly corresponding to scores for the two classes.

We use data from the ChEMBL database \autocites{bentoetal2014}, previously used in the literature for other studies of \ml\ applications to drug discovery \autocites{koutsoukasetal2017}. One set with 60\% of the data is used to train and validate the two classifiers. One set with 20\% is used for the calibration of the probability transducer and evaluation of the classifiers. One further data set with 20\% is here used as fictive \enquote{real data} to illustrate the results of our procedure; we call this the \enquote{demonstration set}.

Note that \emph{the additional demonstration dataset has an illustrative purpose only} for the sake of the present example. In a real design \amp\ evaluation of a set of candidate classifiers, the calibration set will at the same time be the evaluation test set, and no further data subset will be necessary, as explained in \sect~\ref{sec:calibration_data}.

In all data sets, class~$0$ (\enquote{inactive}) occurs with a 91\% relative frequency, and class~$1$ (\enquote{active}) with 9\%; a high class imbalance.

Technical details about the setup and training of the two classifiers and of the calculation of the probability-transducer parameters are given in appendix~\ref{sec:MCMC}.

\subsection{Probability-transducer curves}
\label{sec:demo_curves}

Mathematical details about the expression of the joint probability $p(c,y)$, \eqn~\eqref{eq:representation_P}, for the \RF\ and the \CNN\ are given in appendix~\ref{sec:MCMC}.

\subsubsection{Random forest}
\label{sec:curve_RF}

% The joint probability of class $c$ and output $y$, \eqn~\eqref{eq:representation_P}, for the \RF\ is expressed by the sum
% \begin{equation}
%   \label{eq:P_c_out_RF}
%   p(c, y) = \sum_{k} q_{k}\
%   [c\ \alpha_{k} + (1-c)\ (1-\alpha_{k})]\ 
%   \No(y \| \mu_{k}, \sigma_{k})
% \end{equation}
% where $\No(\cdot)$ is a Gaussian as in \eqn~\eqref{eq:distr_output}. The sum contains $2^{18} \approx 260\,000$ terms.
% % ; $L(y)$ is a logit transformation composed with a small contraction around $0.5$, to take care of $0$ and $1$ values, which maps the output $y\in \clcl{0,1}$ onto the real numbers \autocites[\cf][\sect~3.3 \eqn~(19)]{johnson1949}; $L'(y)$ is its Jacobian determinant;

\begin{figure}[!t]
% P range:  0.1411262 92.9317592
  \centering
  \includegraphics[width=\linewidth]{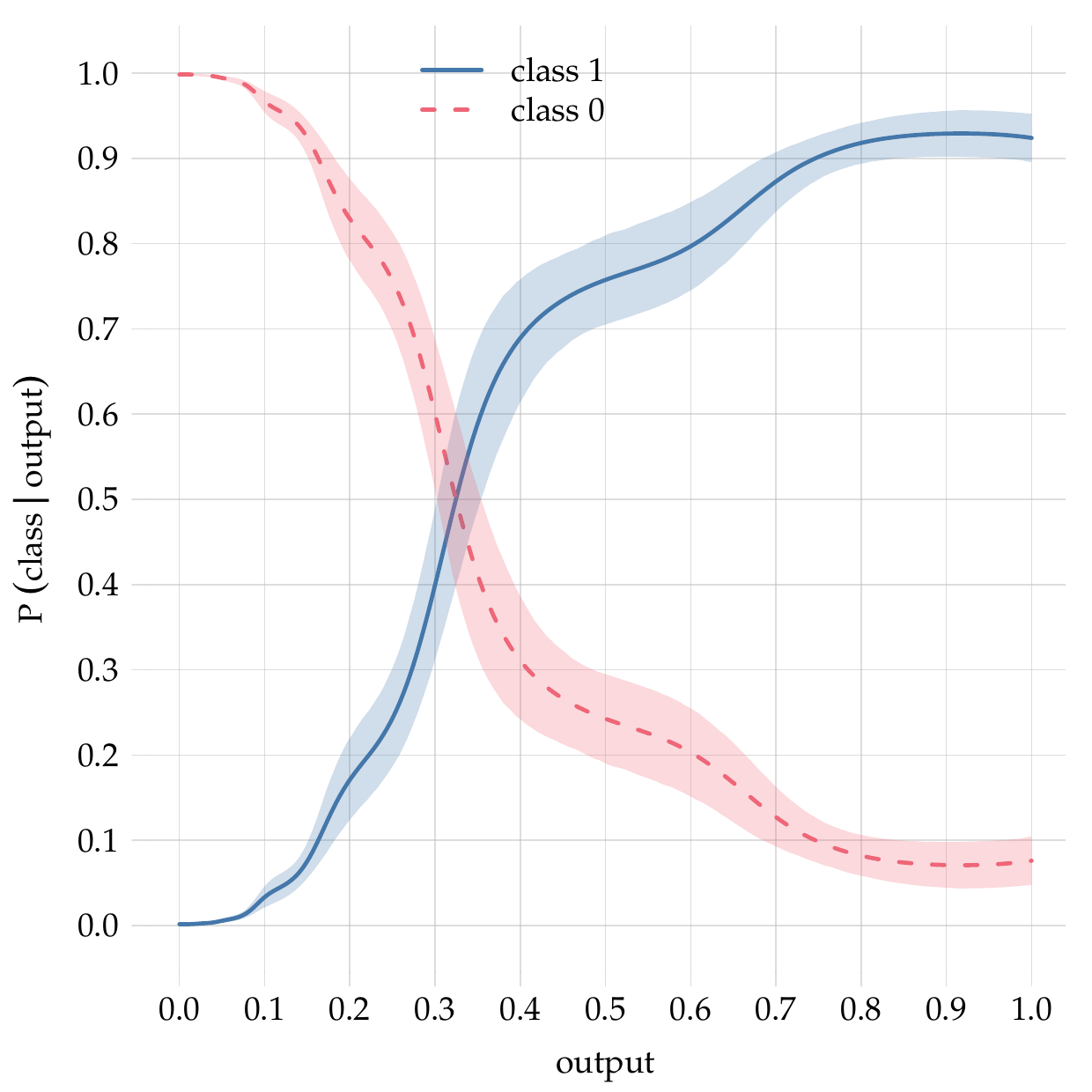}\\
  \caption{Probabilities of class~1 (\enquote{active}, \textcolor{blue}{blue solid curve}) and class~0 (\enquote{inactive}, \textcolor{red}{red dashed curve}) conditional on the \rf\ output. Their extremal values are 0.0014 and 0.929. The \rf\ output clearly cannot be interpreted as a probability. The shaded region around each curve represents its 12.5\%--87.5\% range of possible variability if more data were used to calculate the probabilities.}
  \label{fig:prob_curve_RF}
\end{figure}
Figure~\ref{fig:prob_curve_RF} shows the probabilities of classes~1 and~0 conditional on the \rf\ output: $\p(\texts{class 1} \| \texts{output})$ and $\p(\texts{class 0} \| \texts{output})$. It also shows the range of variability that these probabilities could have if more data were used for the calibration: with a 75\% probability they would remain within the shaded regions. This variability information is provided for free by the calculation; we plan to discuss and use it more in future work.

This curve is not a straight line, which means that the \rf\ output cannot be interpreted as a probability. Note that for an output of around 0.32 both classes are equally probable. For an output of 0.5, class~1 has a probability of around 75\%. 

The probabilities increase (class~1) or decrease (class~0) monotonically up to output values of around 0.9. The minimum and maximum probabilities are 0.14\% and 92.9\%; these values will be important for a later discussion. The output, if interpreted as a probability for class~1 (\enquote{active}), tends to be too pessimistic for this class (and too optimistic for the other) in a range from roughly 0.25 to 0.95; and too optimistic outside this range. For instance, for an output of 0.3 the probability for class~1 is 40\%; for an output of 1 the probability for class 1 is 92\%.

\subsubsection{Convolutional neural network}
\label{sec:curve_CNN}

% The joint probability of class $c$ and the bivariate output $y\equiv (y_{0}, y_{1})$ of the \CNN\ is expressed by the sum
% \begin{multline}
%   \label{eq:P_c_out_CNN}
%   p(c, y_{0}, y_{1}) ={}\\ \sum_{k} q_{k}\
%   [c\ \alpha_{k} + (1-c)\ (1-\alpha_{k})]\
%   \No(y_{0} \| \mu_{0k}, \sigma_{0k})\ 
%   \No(y_{1} \| \mu_{1k}, \sigma_{1k})
% \end{multline}
% containing again $2^{18} \approx 260\,000$ terms, and with parameters analogous to those of \eqn~\eqref{eq:P_c_out_RF}.

\begin{figure}[!t]
  \centering
% P range: 0.1363446 92.3322319
\includegraphics[width=\linewidth]{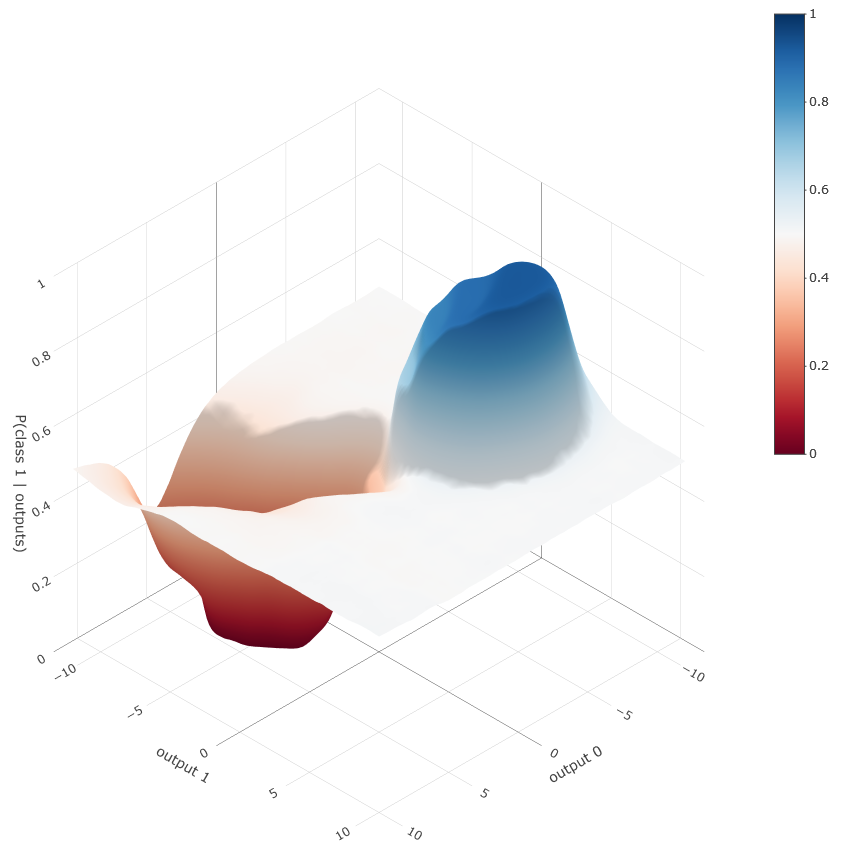}\\[4em]%
%\hfill\makebox[0pt][r]{\smash{\includegraphics[width=0.4\linewidth]{softmaxr_surface_CNN_transp.png}}}\\
% \vspace{-0.75\linewidth}
% \hspace{\stretch{1}}
% \makebox[0pt]{\smash{\includegraphics[width=0.33\linewidth]{softmaxr_surface_CNN.png}}}\hspace{\stretch{3}}\mbox{}
% \\
% \vspace{0.65\linewidth}
\parbox[b]{0.63\linewidth}{\caption{Probability of class~1 conditional on the \cnn\ outputs. Its extremal values are 0.0014 and 0.923. Plot on the side: softmax function of the same outputs, for comparison.\label{fig:prob_curve_CNN}}}\hfill%
\parbox[b]{0.34\linewidth}{\smash{\includegraphics[width=\linewidth]{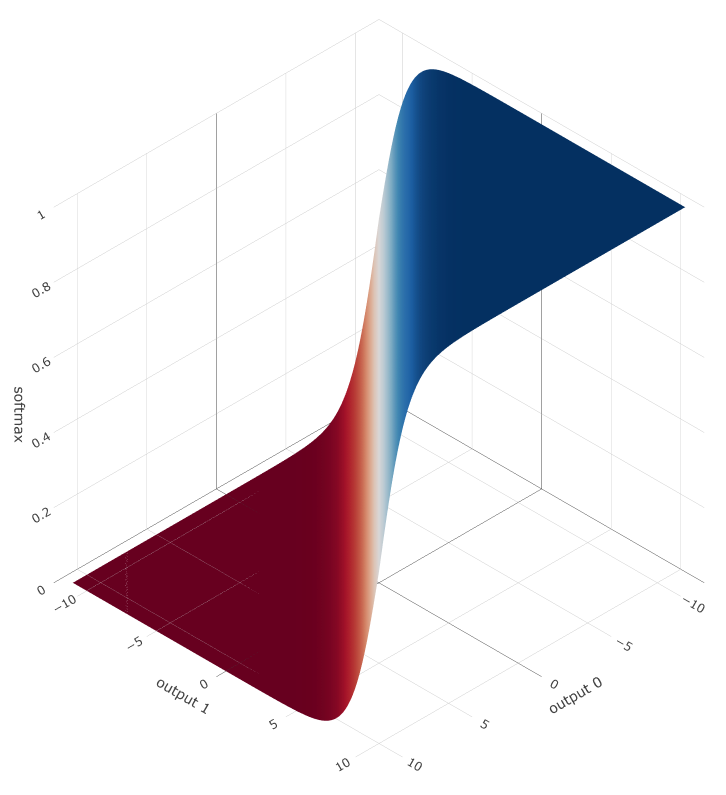}}}
\end{figure}
Figure~\ref{fig:prob_curve_CNN} shows the probability of class~1 conditional on the bivariate output of the \CNN, $\p(\texts{class 1} \| \texts{outputs})$. Its extremal values are 0.14\% and 92.3\%. It is interesting to compare this probability with the softmax function of the outputs, shown in the smaller side plot, typically used as a proxy for the probability.
% \begin{figure}[!t]
%   \centering
% \includegraphics[width=0.66\linewidth]{softmaxr_surface_CNN.png}\\
%   \caption{Softmax function of the \cnn\ outputs}
%   \label{fig:softmax_CNN}
% \end{figure}

\begin{figure}[!t]
  \centering
  \includegraphics[width=\linewidth]{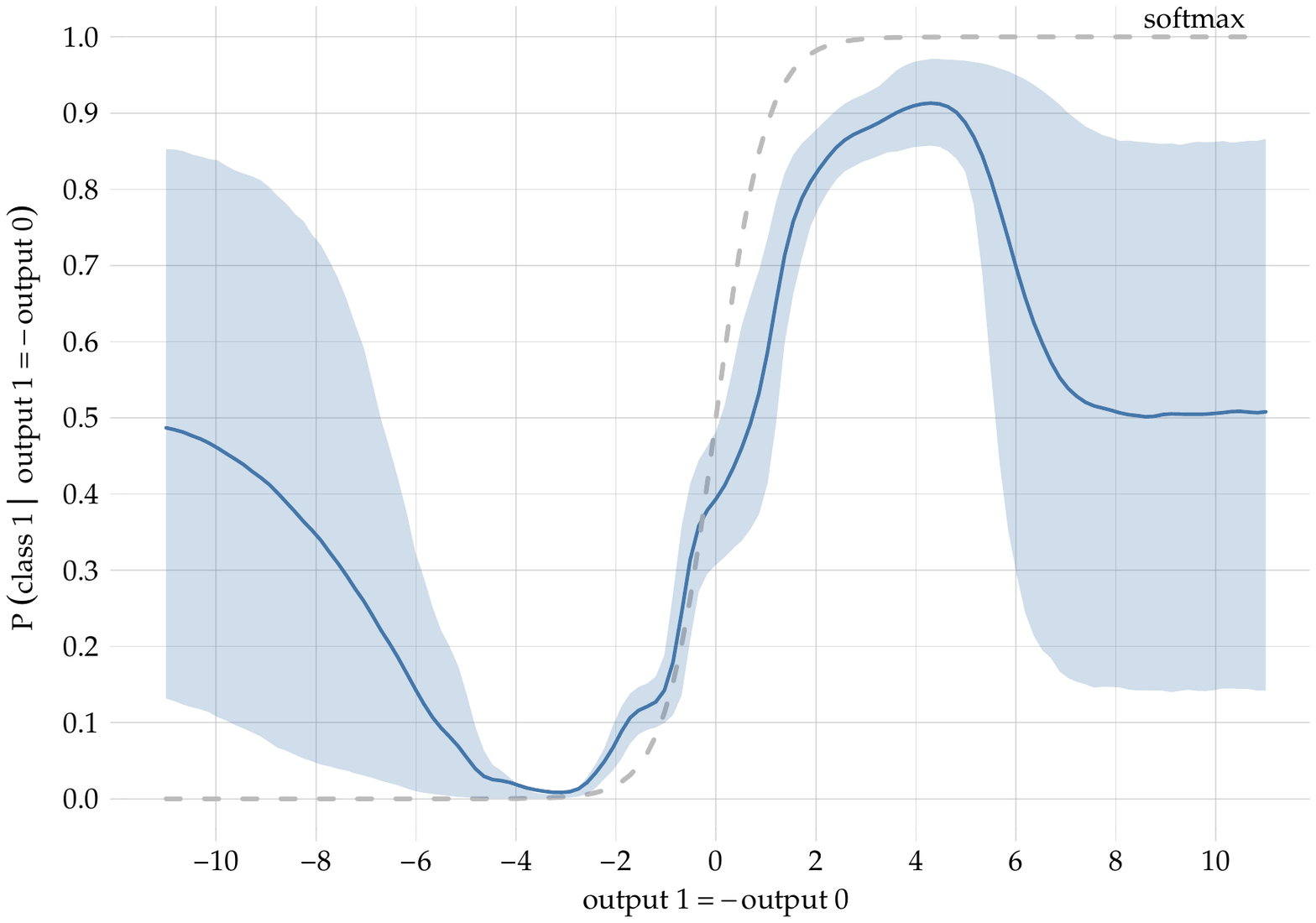}\\
  \caption{Cross-section of the probability surface of \fig~\ref{fig:prob_curve_CNN} across the bisector of the II and IV quadrants of output space. The shaded region represents the 12.5\%--87.5\% range of possible variability upon increase of the calibration dataset. The cross-section of the softmax function (\textcolor{mygray}{grey dashed curve}) is also shown for comparison.}
  \label{fig:prob_curve_CNN_section}
\end{figure}

A cross-section of this probability surface along the bisector of the II and IV quadrants of the output space is shown in \fig~\ref{fig:prob_curve_CNN_section}, together with the cross-section of the softmax. The probability takes on extremal values, around 1\% and 90\%, only in very narrow ranges, and quickly returns and extrapolates to 50\% everywhere else. The softmax, on the other hand, extrapolates to extreme probability values -- a known problem of neural networks \autocites{galetal2016}. The conservative extrapolation of the transducer is also reflected in the 75\% interval of possible variability of the probability (shaded region), which becomes extremely wide at the extremities.

\subsection{Results on demonstration data}
\label{sec:demo_evaluation}

The essential point of the decision-theoretic approach is that we first need to specify the utilities involved in the classification problem, because they determine
\begin{enumerate*}[label=(\roman*)]
\item together with the probabilities, which class we choose in each single instance;
\item the metric to evaluate a classifier's performance.
\end{enumerate*}
The utilities are assembled into a utility matrix which we write in the format
\begin{equation}
\label{eq:format_matrix}
  \rotatebox[origin=c]{90}{
    \clap{\textit{\parbox{4em}{\centering\scriptsize decision\\$1\hspace{1.5em}0$}}
    }}\ 
    \overbracket[0pt]{\umatrix{
    \text{\footnotesize True 0} }{ \text{\footnotesize False 0}}{
    \text{\footnotesize False 1} }{ \text{\footnotesize True 1}}
      }^{
      \clap{\textit{\parbox{6em}{\centering\scriptsize true class\\$0\hspace{3em}1$}}
    }} \ .
\end{equation}
We call \emph{equivalent} two utility matrices that differ by a constant additive term and a positive multiplicative term, since changes in the zero or unit of measurement of utilities do not affect comparative evaluations.

\smallskip

For illustration we choose four utility matrices:
\begin{equation}
  \label{eq:utility_matrices_demo}
  \overbracket[0pt]{\umatrix{1}{0}{0}{1}}^{\text{utility case I}} \qquad
  \overbracket[0pt]{\umatrix{1}{-10}{0}{10}}^{\text{utility case II}} \qquad
  \overbracket[0pt]{\umatrix{1}{0}{-10}{10}}^{\text{utility case III}} \qquad
  \overbracket[0pt]{\umatrix{10}{0}{-10}{1}}^{\text{utility case IV}} \ .
\end{equation}
Case~I represents any case where the correct classification of either class is equally valuable; and the incorrect classification, equally invaluable. Note that this utility matrix is equivalent to any other of the form $\sumatrix{a}{b}{b}{a}$ with $a>b$. Accuracy is the correct metric to evaluate this case. Case~II represents any case where the correct classification of class~1, \enquote{active} or \enquote{positive}, is ten times more valuable than that of class~0, \enquote{inactive} or \enquote{negative}, and its incorrect classification is as damaging as correct classification is valuable. The remaining two cases are interpreted in an analogous way. The \enquote{value} could simply be the final average monetary revenue at the end of the drug-discovery project that typically follows any of these four situations. Of particular relevance to drug discovery, where false positives are known to be especially costly \autocites{sinketal2010,hingoranietal2019}, is the utility matrix of case~II and possibly that of case~III.

We consider each of these utility matrices, in turn, to be the one underlying our classification problem. In each case we perform the classification of every item -- a molecule -- in the demonstration data as follows:
% , first by applying the probability transducer and then by finding the class with maximal expected utility, as explained in \sect~\ref{sec:utility_classification}.
\begin{enumerate}[label=\arabic*.]
\item feed the features of the item to the classifier and record its output
\item feed this output to the probability transducer and record the resulting probability for class~1; form the normalized probability vector for the two classes 
\item multiply the probability vector by the utility matrix to determine the expected utility of each class choice, \eqn~\eqref{eq:expe_utilities}
\item choose the class with higher expected utility (ties have to be decided unsystematically, to avoid biased results), \eqn~\eqref{eq:max_expe_utility}.
\end{enumerate}

\subsubsection{Confusion matrices -- and a peculiar situation}

Once all items in the demonstration dataset are classified, we compare their chosen classes with their true ones and compute the resulting confusion matrix, which we also write in the format~\eqref{eq:format_matrix}. The confusion matrices for all cases, methods, and algorithms are presented in table~\ref{tab:results_CM} on page~\pageref{tab:results_CM}. Ties (both classes were equally preferable) are solved by giving half a point to each class.

The standard method produces the same confusion matrix in all four utility cases because it does not use utilities to choose a class. The augmentation produces instead a different confusion matrix in each utility case: even if the class probabilities for a give datum are the same in all cases, the threshold of acceptance varies so as to always be optimal for the utilities involved.

A peculiar case of this automatic optimization of the threshold is visible for the transducer applied to either algorithm in case~IV: it leads, for both the \RF\ and the \CNN, to the confusion matrix
\begin{equation}
  \label{eq:peculiar_CM}
  \umatrix{3262}{326}{0}{0}
\end{equation}
which means that \emph{all} items were classified as \enquote{0}, \enquote{inactive}. How can this happen? Let us say that the probabilities for class~0 and~1, determined by the transducer from algorithm output, are $1-p$ and $p$. In case~IV, the expected utilities of choosing class~0 or~1 are given by the matrix multiplication $\sumatrix{10}{0}{-10}{1} \begin{bsmallmatrix}1-p\\p\end{bsmallmatrix}$:
\begin{equation}
  \label{eq:expe_utilities_peculiar}
  \begin{aligned}
    \texts{choose 0:\; expect}&& 10 \cdot (1-p) + 0 \cdot p &= 10 - 10\ p \ ,
                                                            \\
    \texts{choose 1:\; expect}&& -10 \cdot (1-p) + 1 \cdot p &= -10 + 11\ p \ .
  \end{aligned}
\end{equation}
It is optimal to choose class~1 only if (disregarding ties)
\begin{equation}
  \label{eq:when_choose_0}
  -10 + 11\ p > 10 - 10\ p
  \qquad\text{or}\qquad
  p > 20/21 \approx 0.952 \ ,
\end{equation}
that is, only if the probability of class~1 is higher than 95\%. The threshold is so high because on the one hand there is a high cost ($-10$) if the true class is not~1, and on the other hand a high reward ($10$) if the true class is indeed~0. Now, a look at the transducer curve for the \RF, \fig~\ref{fig:prob_curve_RF}, shows that the transducer never assigns a probability higher than 93\% to class~1. Similarly the transducer for the \CNN, \fig~\ref{fig:prob_curve_CNN}, never reaches probabilities above 92\%. So the threshold of 95\% will never be met in either case, and no item will be classified as~1. It is simply never rewarding, on average, to do so \autocites[\cf\ the analysis by]{drummondetal2005}. It can be seen that this situation will occur with any utility matrix $\sumatrix{a}{b}{c}{d}$, with $a>c$ and $d>b$, such that $\frac{a-c}{a-c+d-b}\approx 0.93$.

This peculiar situation has a notable practical consequence. We have found the transducer curve for a classifier, and see that its maximum probability for class~1 is 93\%. We have assessed that the utilities involved are $\sumatrix{10}{0}{-10}{1}$, so the threshold to classify as~1 is 95\%. Then we immediately find that \emph{there is no need to employ that classifier, in this utility case}: it is simply more profitable to automatically treat all data as class~0.

\medskip

A look at the other confusion matrices of table~\ref{tab:results_CM} shows the effect of the automatic threshold optimization also in utility cases~II and~III: as the cost of some misclassification increases, the number of the correspoding misclassifications decreases.

\subsubsection{Utility yields}

We can finally assess the performance of both classifiers, with and without augmentation, on the demonstration dataset. As explained in our companion work \autocites{dyrlandetal2022} and summarized in \sect~\ref{sec:utility_classification}, the correct metric for such performance must naturally depend on the utilities that underlie the problem. It is the utility yield per datum produced by the classifier on the dataset, obtained by taking the grand sum of the products of the homologous elements of the utility matrix $(U_{ij})$ and the confusion matrix $(C_{ij})$:
\begin{equation}
  \label{eq:performance_metric}
  \sum_{ij} U_{ij}\ C_{ij} \ .
\end{equation}

\begin{table}[!p]
  \centering
  \small
  \begin{tabular*}{\linewidth}{clw{c}{0.12\linewidth}w{c}{0.12\linewidth}w{c}{0.12\linewidth}w{c}{0.12\linewidth}}
&    &$\sumatrix{1}{0}{0}{1}$
 &$\sumatrix{1}{-10}{0}{10}$
 &$\sumatrix{1}{0}{-10}{10}$
 &$\sumatrix{10}{0}{-10}{1}$
 % &$\sumatrix{100}{0}{-100}{1}$
    \\[4\jot]
% ## standard   0.968   1.364    5.553   8.954   88.92
% ## transducer 0.974   1.720    9.632   9.091   90.91
% ##            0.620  26.100   73.460   1.530    2.24
 &\parbox{0.21\linewidth}{\color{red}\rlap{standard method}}
 &\multicolumn{4}{c}{\textcolor{red}{$\sumatrix{3225}{79.5}{37}{246.5}$}}
 \\[2\jot]
 \smash{\rotatebox[origin=l]{90}{\hspace{1.5em}\clap{\parbox{\widthof{Random}}{\footnotesize Random\\Forest}}}}
 &\parbox{0.21\linewidth}{\color{blue}augmentation}
 &\textcolor{blue}{$\sumatrix{3207}{38}{55}{288}$}
 &\textcolor{blue}{$\sumatrix{3050}{7}{212}{319}$}
 &\textcolor{blue}{$\sumatrix{3207}{40}{55}{286}$}
 &\textcolor{blue}{$\sumatrix{3262}{326}{0}{0}$}
 %&\textcolor{blue}{$\sumatrix{3262}{326}{0}{0}$}
 \\[6\jot]
%%%%%%%%
% ## standard   0.959   1.518    7.237   8.628   85.58
% ## transducer 0.962   1.645    9.591   9.091   90.91
% ##            0.310   8.370   32.530   5.370    6.23
 &\parbox{0.21\linewidth}{\color{red}\rlap{standard method}}
 &\multicolumn{4}{c}{\textcolor{red}{$\sumatrix{3165}{49}{97}{277}$}}
 \\[2\jot]
 \smash{\rotatebox[origin=l]{90}{\hspace{1.5em}\clap{\parbox{\widthof{Network}}{\footnotesize Neural\\Network}}}}
 &\parbox{0.21\linewidth}{\color{blue} augmentation}
 &\textcolor{blue}{$\sumatrix{3189}{65}{73}{261}$}
 &\textcolor{blue}{$\sumatrix{2882}{12}{380}{314}$}
 &\textcolor{blue}{$\sumatrix{3189}{66}{73}{260}$}
 &\textcolor{blue}{$\sumatrix{3262}{326}{0}{0}$}
 %&\textcolor{blue}{$\sumatrix{3262}{326}{0}{0}$}
 \end{tabular*}
  \\[1em]
  \caption{Confusion matrices from demonstration dataset}
  \label{tab:results_CM}

\vspace{6em}
  
% classes
% 3262  326 
%
% baserates
% 0.90914158 0.09085842 
% \begin{table}[!t]
%   \centering
%   \small
  \begin{tabular*}{\linewidth}{clw{c}{0.12\linewidth}w{c}{0.12\linewidth}w{c}{0.12\linewidth}w{c}{0.12\linewidth}}
&    &$\sumatrix{1}{0}{0}{1}$
 &$\sumatrix{1}{-10}{0}{10}$
 &$\sumatrix{1}{0}{-10}{10}$
 &$\sumatrix{10}{0}{-10}{1}$
 % &$\sumatrix{100}{0}{-100}{1}$
     \\[2\jot]
 &\parbox{0.21\linewidth}{\rlap{\color{mygray}\footnotesize min achievable utility}}
% ## >            [,1]
% ## [1,]  0.0000000
% ## [2,] -0.9085842
% ## [3,] -9.0914158
% ## [4,] -9.0914158
% ## [5,] -9.0914158
 &\textcolor{mygray}{\footnotesize $0$}
 &\textcolor{mygray}{\footnotesize $-0.91$}
 &\textcolor{mygray}{\footnotesize $-9.09$}
 &\textcolor{mygray}{\footnotesize $-9.09$}
 %&\textcolor{mygray}{\footnotesize $-90.9$}
     \\[0\jot]
 &\parbox{0.21\linewidth}{\rlap{\color{mygray}\footnotesize max achievable utility}}
% ## [1,] 1.000000
% ## [2,] 1.817726
% ## [3,] 1.817726
% ## [4,] 9.182274
% ## [5,] 1.817726
 &\textcolor{mygray}{\footnotesize 1}
 &\textcolor{mygray}{\footnotesize 1.82}
 &\textcolor{mygray}{\footnotesize 1.82}
 &\textcolor{mygray}{\footnotesize 9.18}
 %&\textcolor{mygray}{\footnotesize 91.0}
    \\[5\jot]
% ## +                 [,1]     [,2]     [,3]     [,4]     [,5]
% ## standard   0.9675307 1.364270 1.482720 8.953874 1.482720
% ## transducer 0.9740803 1.719621 1.537625 9.091416 1.537625
 &\parbox{0.21\linewidth}{\color{red}\rlap{standard method}}
 &\textcolor{red}{0.968}
 &\textcolor{red}{1.36}
 &\textcolor{red}{1.48}
 &\textcolor{red}{8.95}
 %& \textcolor{red}{88.9}
 \\[1\jot]
  \smash{\rotatebox[origin=l]{90}{\hspace{1.5em}\clap{\parbox{\widthof{Random}}{\footnotesize Random\\Forest}}}}
 &\parbox{0.21\linewidth}{\color{blue}\bfseries augmentation}
 &\textcolor{blue}{\bfseries 0.974}
 &\textcolor{blue}{\bfseries 1.72}
 &\textcolor{blue}{\bfseries 1.54}
 &\textcolor{blue}{\bfseries 9.09}
 %&\textcolor{blue}{\bfseries 90.9}
 % \\[2\jot]
 % &\parbox{0.21\linewidth}{\footnotesize relative increase}
 % &\footnotesize+0.6\%
 % &\footnotesize+26\%
 % &\footnotesize+73\%
 % &\footnotesize+1.5\%
 % % &\footnotesize+2.2\%
 \\[5\jot]
%%%%%%%%
%  &\parbox{0.21\linewidth}{\clap{\color{mygray}\footnotesize min achievable utility}}
% % ##             [,1]
% % ## [1,]   0.0000000
% % ## [2,]  -0.9085842
% % ## [3,]  -9.0858417
% % ## [4,]  -9.0914158
% % ## [5,] -90.9141583
%  &\textcolor{mygray}{\footnotesize 0}
%  &\textcolor{mygray}{\footnotesize $-0.91$}
%  &\textcolor{mygray}{\footnotesize $-9.09$}
%  &\textcolor{mygray}{\footnotesize $-9.09$}
%  %&\textcolor{mygray}{\footnotesize $-90.9$}
%     \\[2\jot]
% ##                 [,1]     [,2]     [,3]     [,4]     [,5]
% ## standard   0.9593088 1.517559 1.383779 8.627926 1.383779
% ## transducer 0.9615385 1.644928 1.409978 9.091416 1.409978
 &\parbox{0.21\linewidth}{\color{red}\rlap{standard method}}
 &\textcolor{red}{0.959}
 &\textcolor{red}{1.52}
 &\textcolor{red}{1.38}
 &\textcolor{red}{8.63}
 %&\textcolor{red}{85.6}
 \\[1\jot]
 \smash{\rotatebox[origin=l]{90}{\hspace{1em}\clap{\parbox{\widthof{Network}}{\footnotesize Neural\\Network}}}}
 &\parbox{0.21\linewidth}{\color{blue}\bfseries augmentation}
 &\textcolor{blue}{\bfseries 0.962}
 &\textcolor{blue}{\bfseries 1.64}
 &\textcolor{blue}{\bfseries 1.41}
 &\textcolor{blue}{\bfseries 9.09}
 %&\textcolor{blue}{\bfseries 90.9}
%   \end{tabular*}
%   \\[1em]
%   \caption{Utility yields from demonstration dataset}
%   \label{tab:results_utilities}
    \\[2em]
&&\multicolumn{4}{c}{\scriptsize \rule[0.5ex]{4em}{0.25pt}\quad\emph{Rescaled} utility yields, \eqn~\eqref{eq:norm_utility_formula}\quad\rule[0.5ex]{4em}{0.25pt}}
\\[2\jot]
% \vspace{3em}
% % classes
% % 3262  326 
% %
% % baserates
% % 0.90914158 0.09085842 
% % \begin{table}[!t]
% %   \centering
% %   \small
%   \begin{tabular*}{\linewidth}{clw{c}{0.12\linewidth}w{c}{0.12\linewidth}w{c}{0.12\linewidth}w{c}{0.12\linewidth}}
% ##                 [,1]      [,2]      [,3]      [,4]      [,5]
% ## standard   0.9675307 0.8336741 0.9692913 0.9875011 0.9692913
% ## transducer 0.9740803 0.9640155 0.9743243 0.9950279 0.9743243
 &\parbox{0.21\linewidth}{\color{red}\rlap{standard method}}
 &\textcolor{red}{0.968}
 &\textcolor{red}{0.834}
 &\textcolor{red}{0.969}
 &\textcolor{red}{0.988}
 %& \textcolor{red}{0.989}
 \\[1\jot]
 \smash{\rotatebox[origin=l]{90}{\hspace{1.5em}\clap{\parbox{\widthof{Random}}{\footnotesize Random\\Forest}}}}
 &\parbox{0.21\linewidth}{\color{blue}\bfseries augmentation}
 &\textcolor{blue}{\bfseries 0.974}
 &\textcolor{blue}{\bfseries 0.964}
 &\textcolor{blue}{\bfseries 0.974}
 &\textcolor{blue}{\bfseries 0.995}
 %&\textcolor{blue}{\bfseries 1.000}
 \\[5\jot]
%%%%%%%%
%%% normalized scores CNN
% ##                 [,1]      [,2]      [,3]      [,4]      [,5]
% ## standard   0.9593088 0.8898998 0.9602218 0.9696642 0.9602218
% ## transducer 0.9615385 0.9366183 0.9626233 0.9950279 0.9626233
 &\parbox{0.21\linewidth}{\color{red}\rlap{standard method}}
 &\textcolor{red}{0.959}
 &\textcolor{red}{0.890}
 &\textcolor{red}{0.960}
 &\textcolor{red}{0.970}
 %&\textcolor{red}{0.970}
 \\[1\jot]
 \smash{\rotatebox[origin=l]{90}{\hspace{1em}\clap{\parbox{\widthof{Network}}{\footnotesize Neural\\Network}}}}
 &\parbox{0.21\linewidth}{\color{blue}\bfseries augmentation}
 &\textcolor{blue}{\bfseries 0.962}
 &\textcolor{blue}{\bfseries 0.937}
 &\textcolor{blue}{\bfseries 0.963}
 &\textcolor{blue}{\bfseries 0.995}
 %&\textcolor{blue}{\bfseries 1.000}
  \end{tabular*}
  \\[1em]
  \caption{Utility yields from demonstration dataset}
  \label{tab:results_utilities}

\end{table}

The utility yields for the different cases, classifiers, and methods are presented in table~\ref{tab:results_utilities}. The maximum and minimum theoretically achievable yields, which are obtained when all data are correctly classified or incorrectly misclassified, are also shown for each case. Since the maximum and minimum differ from case to case, the table also reports the \emph{rescaled} utilities: for each case, the rescaled utility is obtained by a change in the zero and scale of its measurement unit such that the minimum and maximum achievable yields become $0$ and $1$:
\begin{equation}
  \label{eq:norm_utility_formula}
  \texts{rescaled utility} =
  \frac{\texts{utility} - \texts{theoretical min}}{
    \texts{theoretical max} - \texts{theoretical min}
  } \ .
\end{equation}

\medskip

Let us first compare the two algorithms when employed in the standard way (red). Their performance is very close to the theoretical maximum in most cases, the worse  being the \RF\ in case~II.  The \RF\ outperforms the \CNN\ in  cases~I, III, IV.

Then let us look at the performances obtained with the augmentation (blue bold). We note the following:
\begin{itemize}
\item The augmentation improves the performance of each algorithm in all cases. The improvement also occurs in case~IV for the \RF, where the standard method already had an extremely high performance (rescaled utility of 0.988).
\item In cases~II and~IV the augmentation improves the originally worse algorithm above the originally better one.
\item In case~IV the augmentation brings the utility yield to above 99\% of the theoretical maximum; remember from the previous section that this is achieved by classifying all data as class~0.
\item With augmentation, the \RF\ outperforms the \CNN\ in all cases, with a possible tie for case~IV.
\end{itemize}

\medskip

These were four particular cases only, though. Does the augmentation lead to an improvement (or at least to no change), on average, over all possible utility matrices? We expect this to be the case, owing to the internal consistency of decision theory.

\begin{figure}[t]
  \centering
  \includegraphics[width=0.5\linewidth]{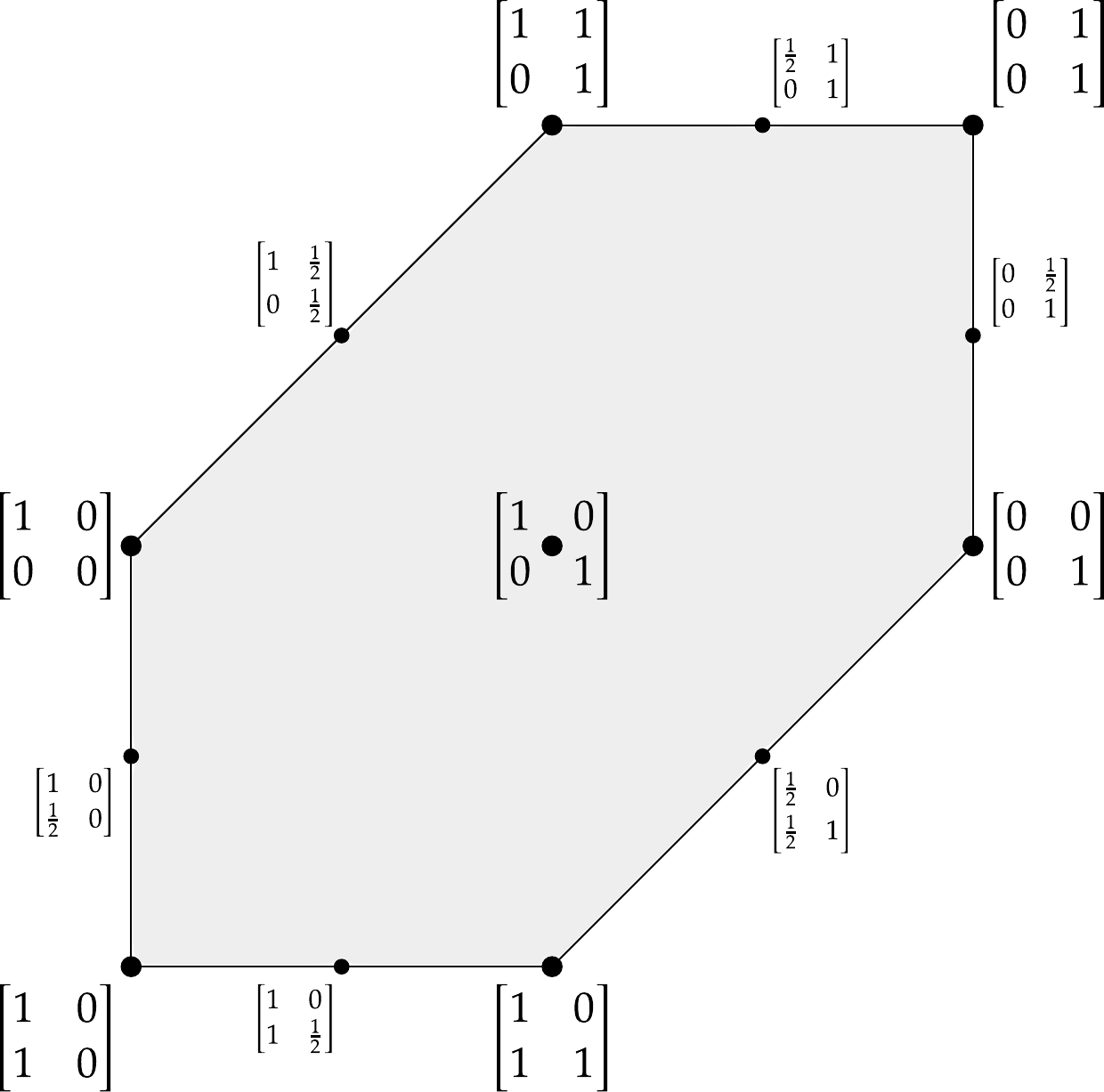}\\
  \caption{Space of utility matrices (modulo equivalence) for binary classification.}
  \label{fig:space_UM}
\end{figure}

We give evidence of this fact by considering a large number (10\,000) of utility matrices selected uniformly from the utility-matrix space (more precisely: manifold) for binary classification. This two-dimensional space, shown in \fig~\ref{fig:space_UM}, is discussed in our companion work \autocites[\sect~3.2]{dyrlandetal2022}.

\begin{figure}[t]
  \centering
  \includegraphics[width=\linewidth]{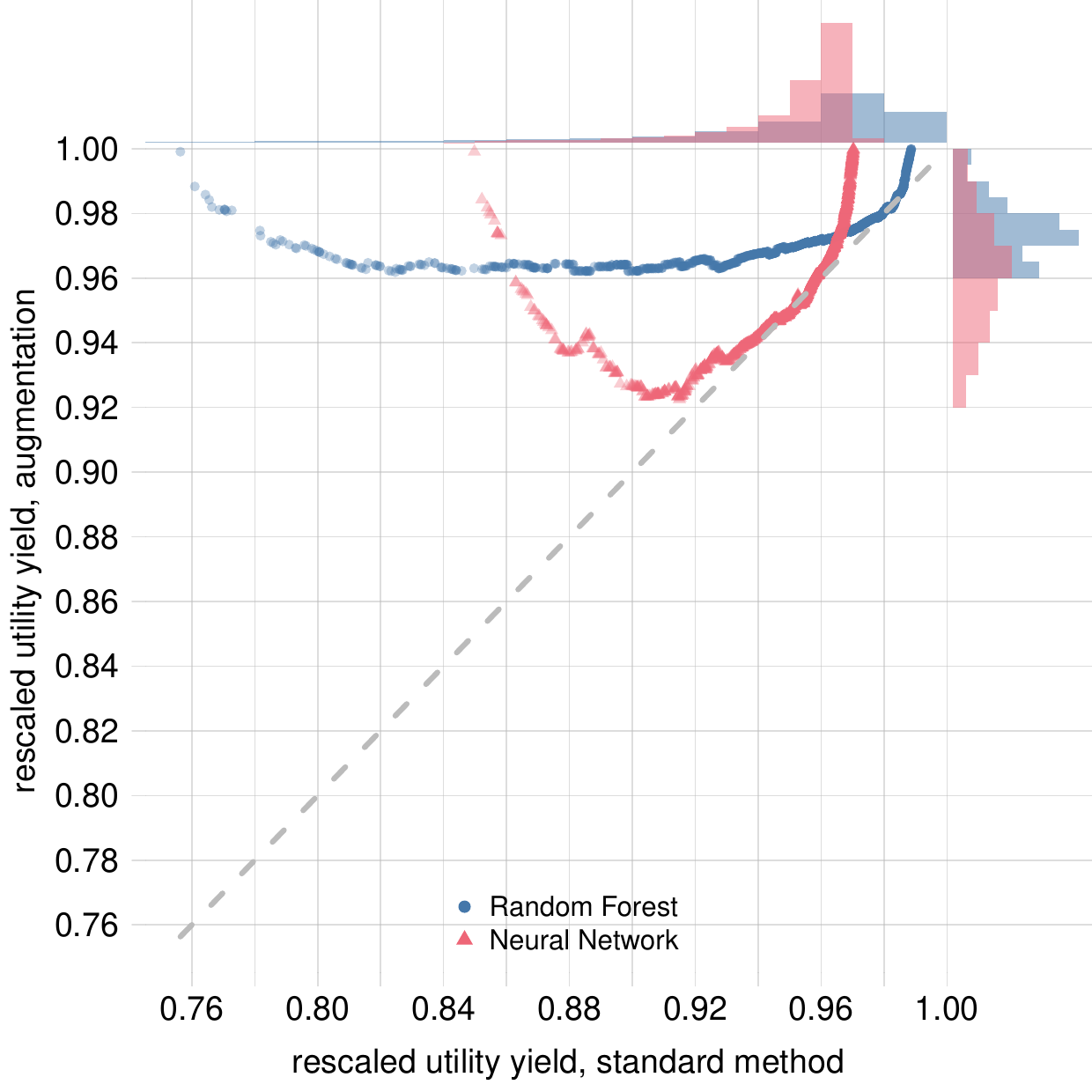}\\
  \caption{Rescaled utility yields obtained using the two classifiers in the standard way, vs those obtained with augmentation, for a uniform distribution of possible utility matrices over the utility-matrix space of \fig~\ref{fig:space_UM}. The augmentation always leads to an improved utility yield, especially in cases where the standard method has a low or high performance. Owing to noise coming from numerical rounding, in some cases the yield from augmentation may appear lower than from the standard method (points below the dashed grey line).}
  \label{fig:RF_gain_UMspace}
\end{figure}

For each of these utility matrices, we calculate the rescaled utility yields obtained by using either classifier in the standard way, and with the augmentation -- probability-transducer \amp\ utility-based classification. The utility yields obtained in the two ways are plotted against each other in \fig~\ref{fig:RF_gain_UMspace}. Histograms of their distributions are also shown on the sides. % of  Only 2\,000 are plotted, in order to show the fact that most of the yields are quite high (they accumulate on the top right corner); but all 10\,000 lie along the jagged curves evident in the plot.

% ## > summary(normallscores[1,])
% ##    Min. 1st Qu.  Median    Mean 3rd Qu.    Max. 
% ##  0.7563  0.9429  0.9670  0.9523  0.9791  0.9887 
% ## > summary(normallscores[3,])
% ##    Min. 1st Qu.  Median    Mean 3rd Qu.    Max. 
% ##  0.9618  0.9682  0.9740  0.9749  0.9796  1.0000 
% ## > summary(CNNnormallscores[1,])
% ##    Min. 1st Qu.  Median    Mean 3rd Qu.    Max. 
% ##  0.8498  0.9466  0.9590  0.9514  0.9653  0.9703 
% ## > summary(CNNnormallscores[3,])
% ##    Min. 1st Qu.  Median    Mean 3rd Qu.    Max. 
% ##  0.9225  0.9478  0.9614  0.9605  0.9733  1.0000 
The augmentation clearly leads to increased utility yields, especially for those cases where the standard performance of the two algorithms is particularly high or low -- compare the left tails of the histograms for the standard method and augmentation. The standard method in some cases has utility yields as low as 0.76 for the \RF\ and 0.85 for the \CNN; whereas the augmentation never leads to yields below 0.96 for the \RF\ and 0.92 for the \CNN. This explains the U-shapes of the scattered points. Note that the minimum values of the plot's axes is 0.75, so the improvement is upon utility yields that are already quite high.

There are a few apparent decreases in the utility yield, in some cases. The extremal relative decreases are $-0.09\%$ for \RF\ and  $-0.2\%$ for \CNN. Given their small magnitude, we believe them to be caused by numerical-precision error rather than to be real decreases % this is actually useful for estimating the magnitude of numerical-rounding errors occurring in the computation and the number of significant digits.

\subsection{From `inactive vs active' to more general decisions}
\label{sec:gen_decision}

In the demonstration just discussed we assumed that the decisions available for each molecule examined were just two: \enquote{molecule is inactive} vs \enquote{molecule is active}, corresponding to the two unknown classes. In a more general drug-discovery problem we could have a different set of decisions, for instance \enquote{discard} vs \enquote{promote to next examination stage} vs \enquote{examine with different method}. Each decision would have its own utilities conditional on the two possible classes, forming a 3\texttimes2 utility matrix. The analysis and calculations of the present section would be easily generalized to such case.

\section{Additional uses of the probability-transducer: an overview}
\label{sec:overview_other_uses}

The probability-transducer presented in \sect~\ref{sec:transducer} and illustrated in the previous section has several other uses and advantages, all of which come for free or almost for free with its calculation. We give a brief overview of them in the present section, leaving a more thorough discussion and applications to future works.

The additional uses are mainly three:
\begin{itemize}
\item Quantification of the possible variability of the transducer probability curve.
\item Evaluation of the optimal algorithm, including the uncertainty about such evaluation.
\item \enquote{Generative use} of the augmented algorithm, even if the original algorithm is not designed for generative use.
\end{itemize}

\subsection{Variability of the transducer's probability curve}
\label{sec:variability_curve}

The output-to-probability function, \eqn~\eqref{eq:prob_output}, such as those plotted in \figs~\ref{fig:prob_curve_RF} and \ref{fig:prob_curve_CNN}--\ref{fig:prob_curve_CNN_section}, is determined by the data in the calibration set. There is the question, then, of how the function could change if we used more calibration data. Such possible variability could be of importance. For example, we may find that the transducer only yields class probabilities around 0.5, and wonder whether this is just a statistical effect of a too small calibration dataset, or whether it would persist even if we used more calibration data.

The calculation of the transducer parameters automatically tells us the probabilities of these possible variations, in the form of  a set of possible alternative transducer curves, from which we can for example calculate quantiles. The shaded regions in \figs~\ref{fig:prob_curve_RF}, \ref{fig:prob_curve_RF_inverse} and \ref{fig:prob_curve_CNN_section} are examples of such probability intervals. Their calculation is sketched in appendix~\ref{sec:variability_prob}.

\subsection{Expected utility of the classifying algorithm}
\label{sec:yield_of_classifier}

At the end of the discussion about the calibration dataset, \sect~\ref{sec:calibration_data}, we gave our assurances that no additional data must be set apart -- with a detrimental reduction in training data -- for evaluation or testing purposes. This is because from the probabilities~\eqref{eq:prob_output}, obtained from the calibration data, we can also calculate \emph{the expected, future utility yield of the augmented algorithm}, once we have specified the utility matrix underlying the particular application. More details about this calculation, which amounts to a low-dimensional integration, are given in appendix~\ref{sec:algorithm_yield}; see especially formula~\eqref{eq:algorithm_utility}.

For the \RF\ and \CNN\ of the demonstration \sect~\ref{sec:demonstration}, for instance, this calculation gives the expected utilities (non-rescaled) of table~\ref{tab:algorithms_utilities}.
% CNN:
%  0.9623081 1.6206641 9.5603046 9.0772883
% RF:
%  0.9725162 1.6815698 9.5864235 9.0783451
\begin{table}[!p]
  \centering
  \begin{tabular}%{\linewidth}%{lw{c}{0.12\linewidth}w{c}{0.12\linewidth}w{c}{0.12\linewidth}w{c}{0.12\linewidth}}
{lcccc}
    &$\sumatrix{1}{0}{0}{1}$
    &$\sumatrix{1}{-10}{0}{10}$
    &$\sumatrix{1}{0}{-10}{10}$
    &$\sumatrix{10}{0}{-10}{1}$
    % &$\sumatrix{100}{0}{-100}{1}$
    \\[2\jot]
    \parbox{0.21\linewidth}{\color{blue}Random Forest}
    &\textcolor{blue}{0.973}
    &\textcolor{blue}{1.68}
    &\textcolor{blue}{9.59}
      &\textcolor{blue}{9.08}
    \\[1\jot]
    \parbox{0.21\linewidth}{\color{red}Neural Net}
    &\textcolor{red}{0.962}
    &\textcolor{red}{1.62}
    &\textcolor{red}{9.56}
      &\textcolor{red}{9.08}
  \end{tabular}
  \caption{Expected utilities for the two algorithms of \sect~\ref{sec:demonstration}}
  \label{tab:algorithms_utilities}
\end{table}
\begin{figure}[!p]
  \centering
  \parbox{0.495\linewidth}{\centering
    \includegraphics[width=\linewidth]{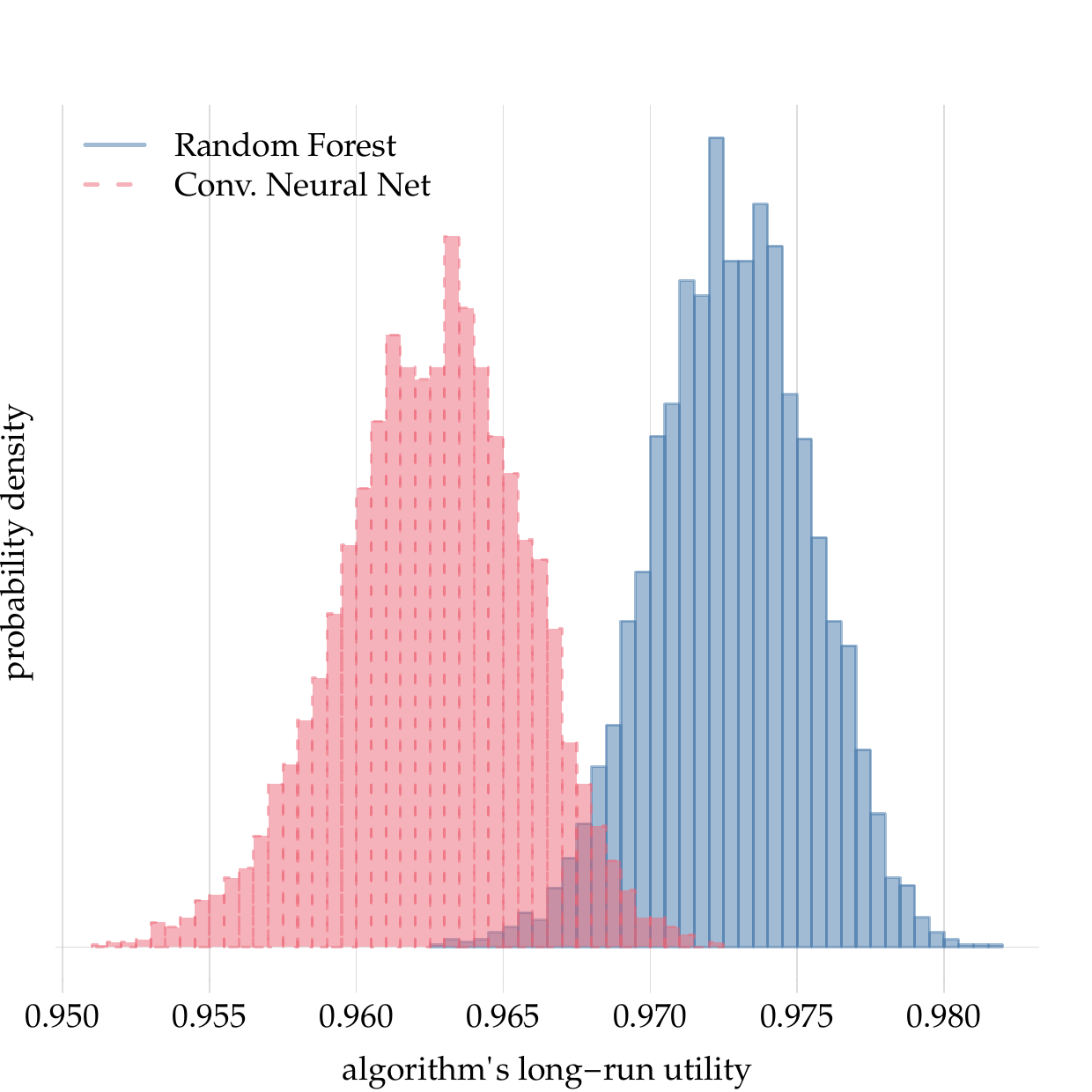}
   \\\footnotesize case I}%
  \hfill
  \parbox{0.495\linewidth}{\centering
    \includegraphics[width=\linewidth]{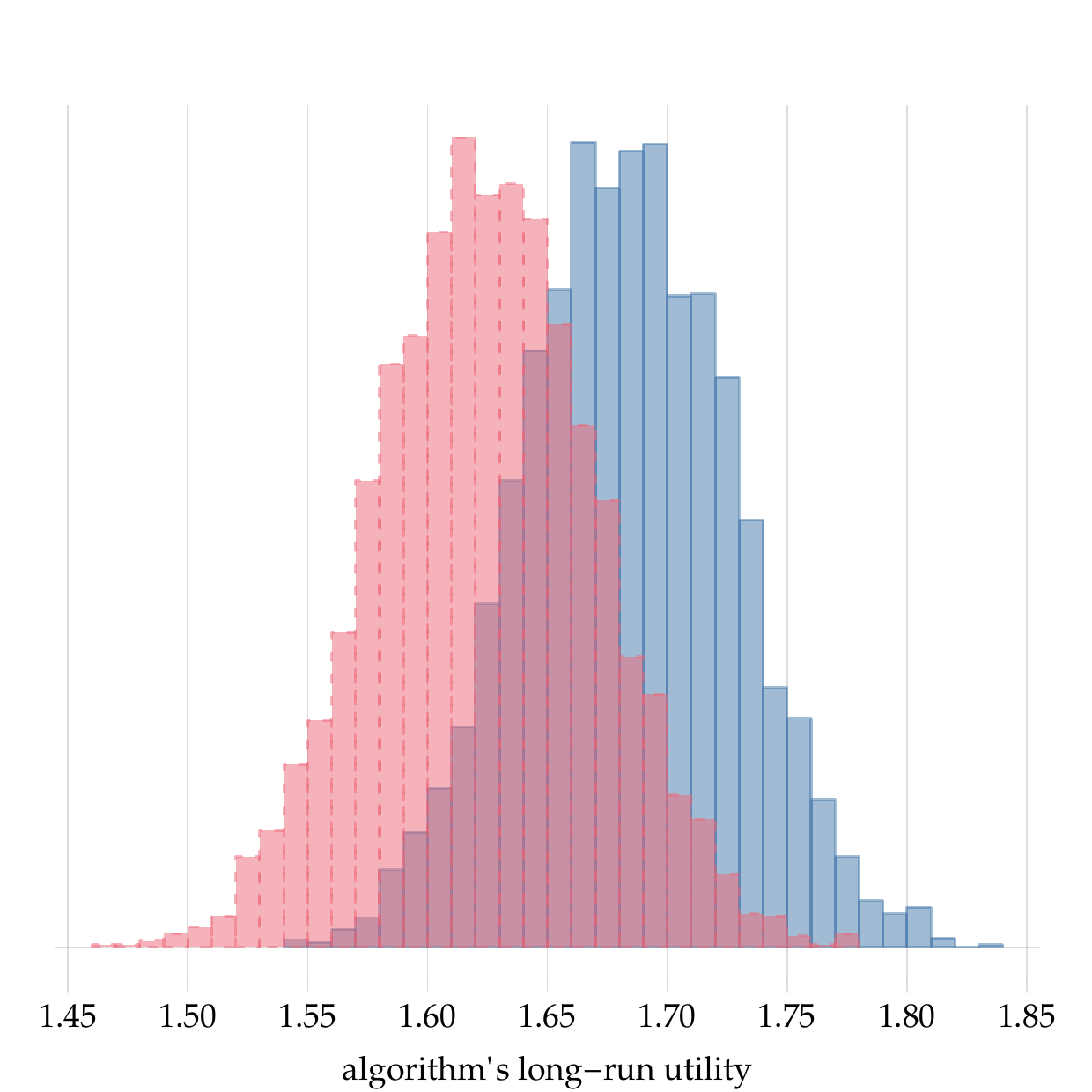}
    \\\footnotesize case II}
  \\[\jot]
  \parbox{0.495\linewidth}{\centering
    \includegraphics[width=\linewidth]{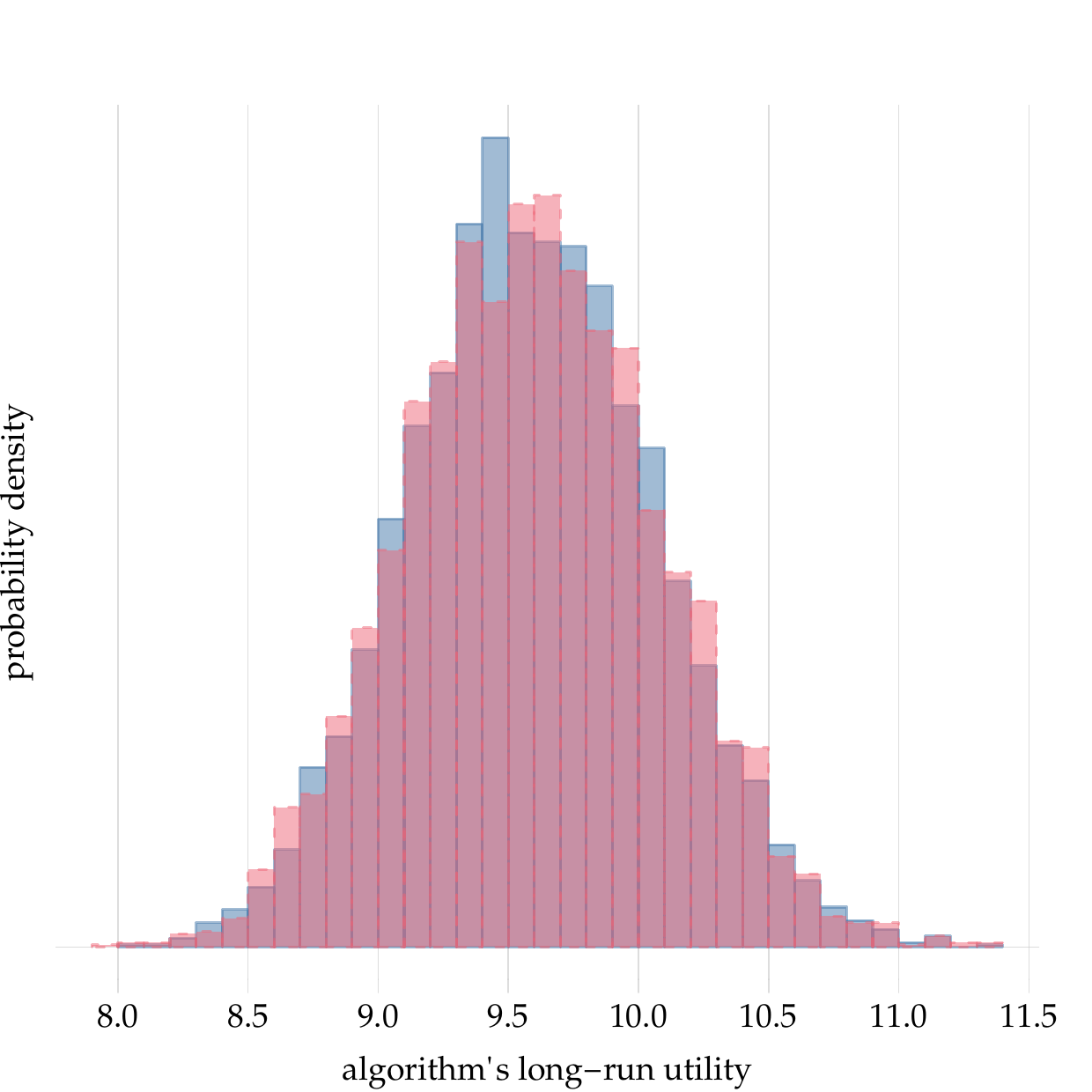}
  \\\footnotesize case III}%
  \hfill
  \parbox{0.495\linewidth}{\centering
    \includegraphics[width=\linewidth]{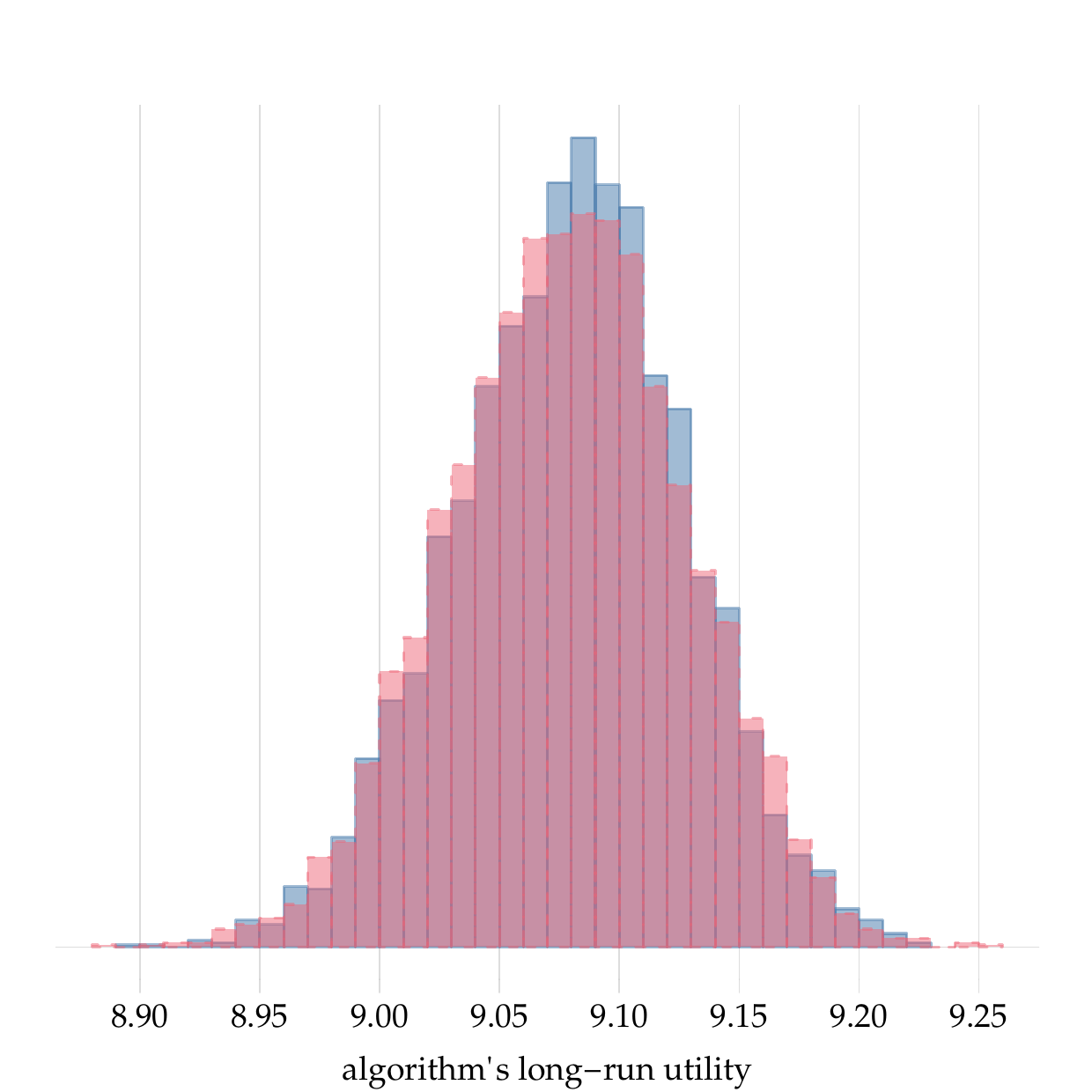}
  \\\footnotesize case IV}
\\
  \caption{Probability distributions of the long-run utility yields of \RF\ and \CNN\ in the four cases of \sect~\ref{sec:demonstration}}
  \label{fig:algorithms_utilities_histograms}
\end{figure}
The augmented \RF\ is expected to be optimal for cases~I and~II, possibly also in case~III, although the difference in utilities is likely affected by numerical-precision error. There is no preference in case~IV.

Let us emphasize again that these values are obtained \emph{from the parameters of the transducer curve, without the need of any additional dataset}. The demonstration dataset discussed in \sect~\ref{sec:demo_overview} was not used for their calculation. The results from that dataset, reported in table~\ref{tab:results_utilities}, corroborate these values.

One may ask: but how can you be sure that what you basically found from the calibration data will generalize to new data? The answer goes back to the discussion at the end of \sect~\ref{sec:calibration_data}, about how the probability calculus works, and to the technical details explained in appendix~\ref{sec:maths_transducer}: the probability calculus automatically considers all possible sets of new data that could be encountered in the future application \autocites[\cf][]{smithetal2006}.

% ## > sapply(1:4,function(i){sum(rfutdistr[i,]>cnnutdistr[i,])/ncol(rfutdistr)})
% ## [1] 0.9934082 0.8276367 0.5039062 0.5075684
In fact, the calculation of an algorithm's expected utility automatically produces a probability distribution of the possible long-run yields the algorithm could give. The distributions for the long-run utilities of the \RF\ and the \CNN\ in cases~I--IV of \sect~\ref{sec:demonstration} are shown in \fig~\ref{fig:algorithms_utilities_histograms}. It can be calculated, \eqn~\eqref{eq:compare_alg_utilities}, that in case~I the \RF\ will very probably, 99\%, be superior to the \CNN. In case~II the probability is somewhat lower, 83\%. In cases~III and~IV it is completely uncertain (50\%) which algorithm will be best.

The evaluation of candidate classifiers' performances and their uncertainties are obviously extremely important for the choice and final deployment of the optimal classifier.

\subsection{Discriminative and generative modes}
\label{sec:effect_transd}

The transducer parameters, calculated as discussed in \sect~\ref{sec:calculation_transducer} and appendix~\ref{sec:maths_transducer}, allow us to calculate not only the \enquote{discriminative} probability of the class given the algorithm's output, formula~\eqref{eq:conditional_c_given_y}, but also the inverse, \enquote{generative} probability \autocites[\sect~21.2.3]{russelletal1995_r2022}[\sect~8.6]{murphy2012} of the output given the class, formula~\eqref{eq:conditional_y_given_c}. The transducer thus allow us to use the original algorithm both in \enquote{discriminative mode} and in \enquote{generative mode}, even if it is not a generative algorithm in itself.

Having an available generative mode is extremely useful, because it is the required way to calculate the class probabilities \emph{if the calibration and training sets do not have the same class frequencies as the real population on which the classifier will be employed}. For instance, two classes may appear in a 50\%/50\% proportion in the calibration set but in a 90\%/10\% proportion in the real population. This discrepancy in the two populations' frequencies can occur for several reasons. Examples: samples of the real population are unavailable or too expensive to be used for calibration purposes; the class statistics of the real population has suddenly changed right after the deployment of the classifier; it is necessary to use the classifier on a slightly different population; or the sampling of calibration data was poorly designed.

In such situations, the discriminative probabilities $\p(c \| y)$ are usually no longer the same in the two populations either, owing to the identity
\begin{equation}
  \label{eq:identity_conditionals}
  \p(c \| y)\ \p(y) \equiv \p(y \| c)\ \p(c) \ .
\end{equation}
Typically, a change in $\p(c)$ leaves $\p(y \| c)$ the same; but then both $\p(y)$ and $\p(c \| y)$ must change as well. This means that the discriminative probabilities the algorithm and transducer have learned from the training and calibration sets are actually wrong: they cannot lead to reliable inferences on the real population.

But, as we just said, the generative probabilities $\p(y \| c)$ often remain the same. And these have been automatically computed in the transducer calibration, formula~\eqref{eq:conditional_y_given_c}.
We can then use them to calculate the probability of class $c$ through Bayes's theorem, by supplying the \emph{population prevalence} $\br_{c}$ of the class:
\begin{equation}
  \label{eq:bayes_baserate}
  p(c \| y,\, \texts{prevalences}) = \frac{p(y \| c)\ \br_{c}}{ \sum_{c} p(y \| c)\ \br_{c} } \ .
\end{equation}
The population prevalences \autocites[\chap~3]{soxetal1988_r2013}[\sect~5.1]{huninketal2001_r2014}, also called \emph{base rates} \autocites{barhillel1980,axelsson2000}, are the relative frequencies of occurrence of the various classes in the population whence our unit originates. This notion is very familiar in medicine and epidemiology. For example, a particular type of tumour can have a prevalence of 0.01\% among people of a given age and sex, meaning that 1 person in 10\,000 among them has that kind of tumour, as obtained through a large survey.

We recommend the outstandingly insightful discussion by \cites{lindleyetal1981} on the problem of population mismatch and on which conditional probabilities to use in that case.

\medskip

\begin{figure}[t]
  \centering
  \includegraphics[width=\linewidth]{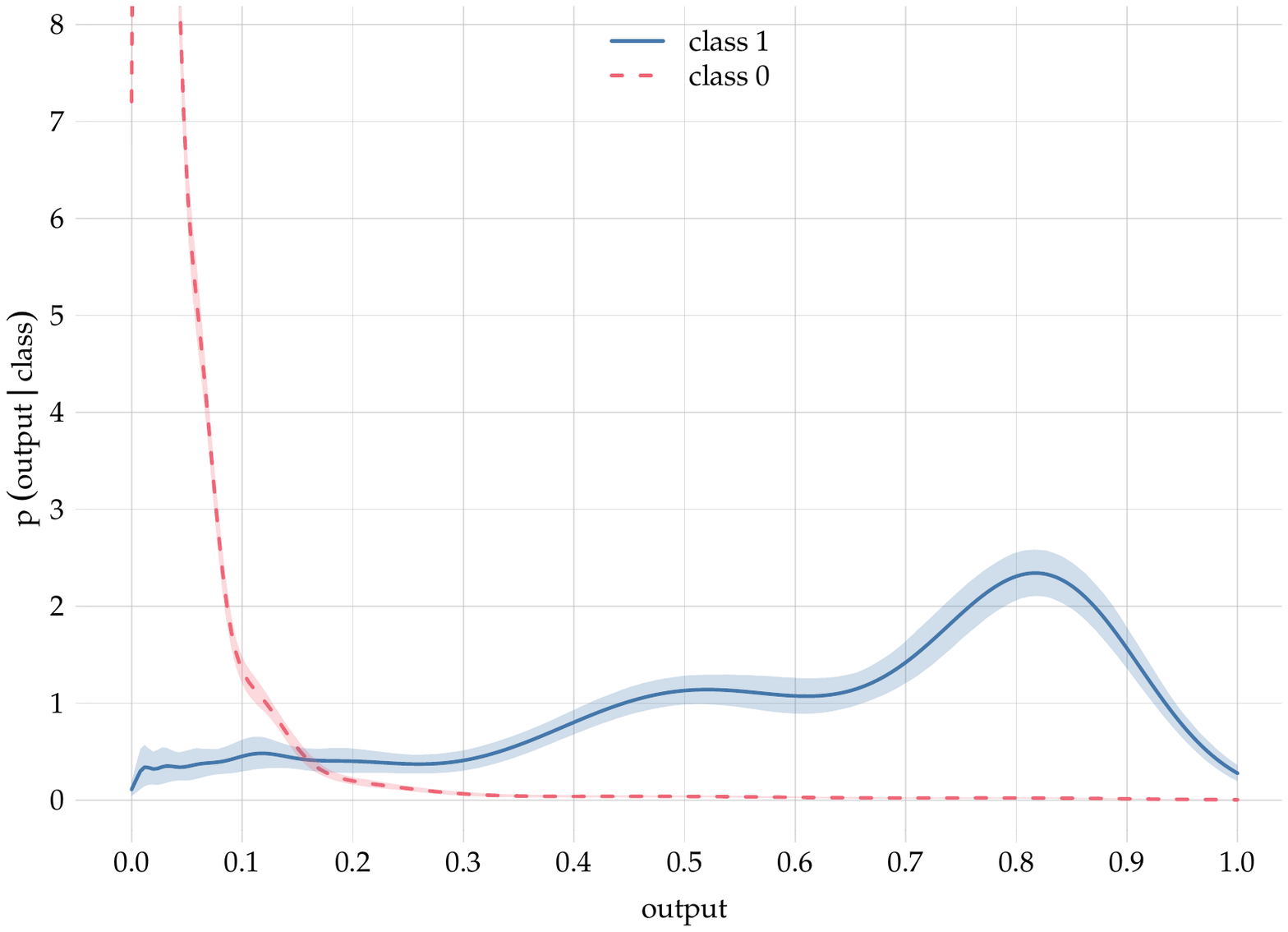}\\
  \caption{Probability densities of the \rf\ output conditional on class~1 (\enquote{active}, \textcolor{blue}{blue solid curve}) and on class~0 (\enquote{inactive}, \textcolor{red}{red dashed curve}, truncated). The shaded region around each curve represents its 12.5\%--87.5\% range of possible variability upon increase of the calibration dataset.}
  \label{fig:prob_curve_RF_inverse}
\end{figure}
Figure~\ref{fig:prob_curve_RF_inverse} shows the \enquote{generative} probability densities $\p(\texts{output} \| \texts{class 1})$, $\p(\texts{output} \| \texts{class 0})$ of the \rf\ output from the demonstration of \sect~\ref{sec:demonstration}. % These probabilities densities are also provided for free by the parameter calculation, as explained in \sect~\ref{sec:calculation_transducer} and appendix. 
The shaded regions are 75\% intervals of possible variability upon increase of the calibration dataset.

There is a high probability of output values close to~0 when the true class is~0 (\enquote{inactive}), and a peak density around 0.8 when the true class is~1 (\enquote{active}), as expected. The density conditional on class~0 is narrower than the one conditional on class~1 owing to the much larger proportion data in the former class. Intuitively speaking, we have seen that most data in class~1 correspond to high output values, but we have seen too few data in this class to reliably conclude, yet, that future data will show the same correspondence. % But there is also a high density of output value~0 when the true class is~1. This peak comes from the fact that the \RF\ had some very low output values

\begin{figure}[!t]
  \centering
  \includegraphics[width=\linewidth]{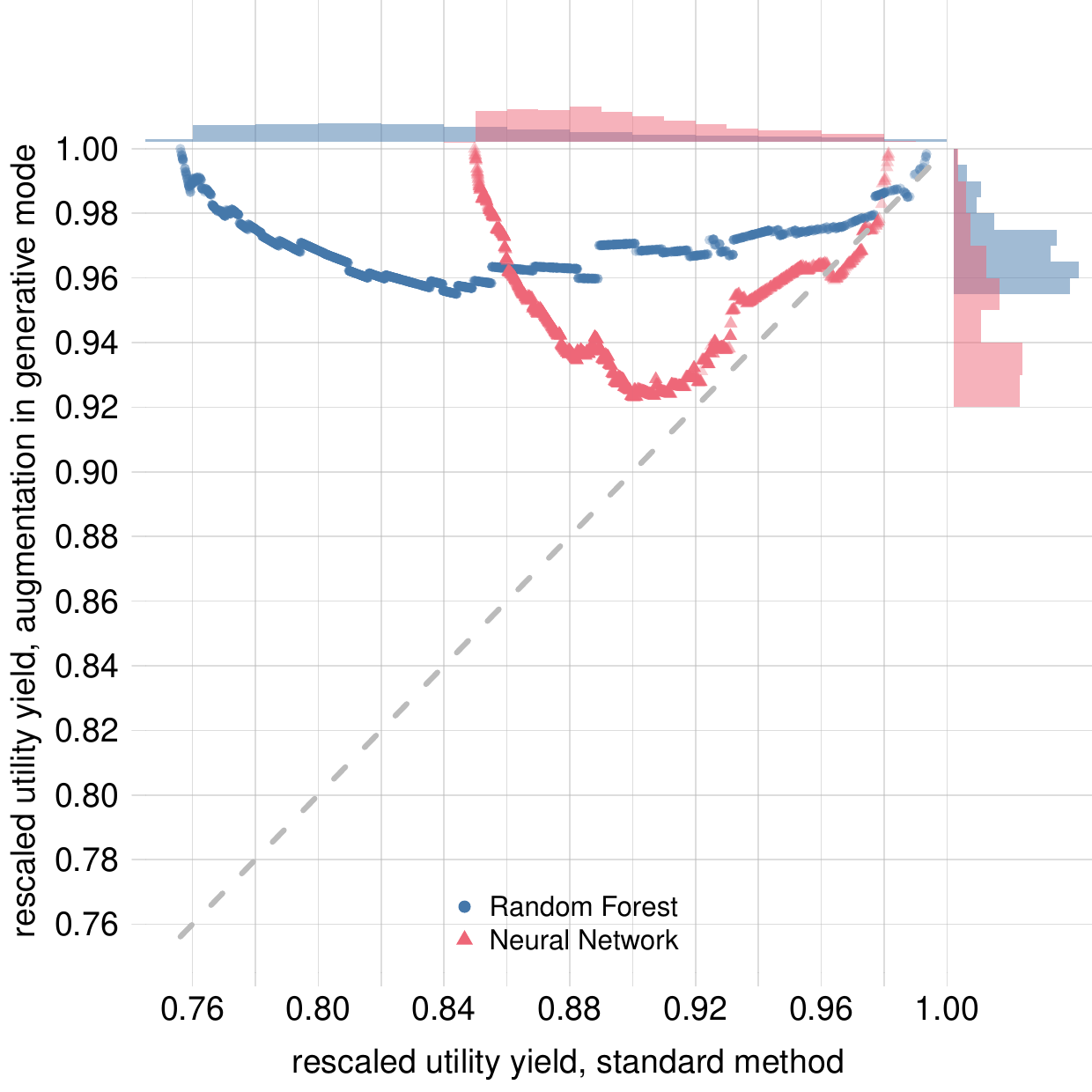}\\
  \caption{Rescaled utility yields obtained using the two classifiers in the standard way, vs those obtained with augmentation in generative mode, on an altered dataset with very different class balance from the training and calibration sets.
The distribution of possible utility matrices is  uniform over their space, as before. The utility yields of the standard method have worsened with respect to those of \fig~\ref{fig:RF_gain_UMspace}, as can be seen from the histogram tails. The augmentation in generative mode, however has not suffered from this dataset mismatch.}
  \label{fig:genRF_gain_UMspace}
\end{figure}

We can show the usefulness of using the probability transducer in generative mode by altering the class frequencies of the demonstration set: we keep all data of class~1 (the less frequent) and unsystematically select a number of data from class~0 equal to half that of class~1. This new demonstration set has thus a proportion 1/3 vs 2/3 of class~0 and class~1: their preponderance has been almost inverted. Finally we apply both classifiers in the standard way and with the augmentation in generative mode, considering again a large number of possible utility matrices, uniformly selected from their space. The rescaled utility yields are shown in \fig~\ref{fig:genRF_gain_UMspace}.
% ## > summary(normallscoresb[1,])
% ##    Min. 1st Qu.  Median    Mean 3rd Qu.    Max. 
% ##  0.7561  0.7964  0.8340  0.8451  0.8844  0.9938 
% ## > summary(normallscoresb[3,])
% ##    Min. 1st Qu.  Median    Mean 3rd Qu.    Max. 
% ##  0.9550  0.9606  0.9673  0.9683  0.9736  1.0000 
% ## > summary(CNNnormallscoresb[1,])
% ##    Min. 1st Qu.  Median    Mean 3rd Qu.    Max. 
% ##  0.8497  0.8721  0.8929  0.8991  0.9209  0.9815 
% ## > summary(CNNnormallscoresb[3,])
% ##    Min. 1st Qu.  Median    Mean 3rd Qu.    Max. 
% ##  0.9213  0.9322  0.9414  0.9467  0.9596  1.0000 

We see that the performance of the standard method has worsened; this is especially manifest by a comparison of the top histograms of \figs~\ref{fig:RF_gain_UMspace} and \ref{fig:genRF_gain_UMspace}. The median utility yield of the \RF\ has gone from 0.967 to 0.834; that of the \CNN\ from 0.959 to 0.893. And yet the augmentation in generative mode is almost unaffected, the median changing from 0.974 to 0.967 for the \RF\ and from 0.961 to 0.941 for the \CNN.

% \mynotez{The discussion below is not true, because the probabilities of the outputs of different classifiers are not independent (they are conditional on the same data)}
% The generative mode is also required if we want to combine the outputs $y_{1}, y_{2}, \dotsc$ of several independent classifiers. According to the probability calculus, the probability of class $c$ is in this case proportional to the probabilities of the outputs \emph{given the class}\autocites[\sect~8.9]{jaynes1994_r2003}{hailperin2006}:
% \begin{equation}
%   \label{eq:combine_probs}
%   \p(c \| y_{1}, y_{2}, \dotsc) \propto \p(c) \times
%   \p(y_{1} \| c) \times \p(y_{2} \| c)\times \dotsm
% \end{equation}
% where $\p(c)$ is a population prevalence or an initial probability assessment.

\section{Summary and discussion}
\label{sec:summary_discussion}

The successful application of \ml\ classifiers in fields such as medicine or drug discovery, which involve high risks and special courses of action, demands that we replace a too-simplistic view of classification with a more articulated and flexible one. A classifier must be able to handle decisions that do not correspond to some unknown classes; it must take into account problem-specific gains and losses arising from such decisions; it must choose not what's likely, but what's optimal; and the uncertainties underlying its operation must be amenable to assessment. And it should preferably face all these requirements with methods based on first-principles guaranteeing consistency and universal applicability.

The basic theory that allows us to face most of these requirements has been around for a long time \autocites[at least since][]{luceetal1957}[\cf][\sect~1.2]{russelletal1995_r2022}: Decision Theory, whose methods keep on see-sawing in machine learning \autocites[\eg][]{selfetal1987,elkan2001,drummondetal2005}. It allows us to consider decisions separate from classes, to evaluate gains and losses, and to decide what's optimal. In the present work we have tried to revive it, showing that its application is straightforward, involves little computational cost, and always leads to improvement on results obtained with standard \ml\ methods, even when these are already nearly optimal.

\medskip

The main obstacle in using decision theory is that it requires proper probabilities, which in many applications might only be obtained at too high computational costs -- \emph{if these probabilities are conditional on the \enquote{features} constituting the input to classifier}.

We have proposed the idea of \emph{using probabilities conditional on the output of the classifier} instead. This is somehow like using the classifier in the guise of a diagnostic test, such as a typical medical test.

This probabilistic quantification is not computationally expensive, can be calculated exactly by Bayesian \emph{model-free} (non-parametric) density-regression methods \autocites{dunsonetal2011}, and \emph{only needs to be done once} per trained algorithm.

We have called the resulting output-to-probability function a \enquote{probability transducer}. Concrete examples are given in \fig~\ref{fig:prob_curve_RF} for the output of a \rf\ classifier, and in \figs~\ref{fig:prob_curve_CNN}, \ref{fig:prob_curve_CNN_section} for the bivariate output of a \CNN. In the latter case, the probability transducer is essentially a replacement of the popular softmax.

The quantification of the probabilistic relationship between a classifier's output and the unknown class requires a \enquote{calibration dataset}, whose role can perfectly be played by the \enquote{test} or \enquote{evaluation} set of standard \ml\ methodology. The calibration dataset also delivers all necessary evaluations; thus a third, additional test set is not required.

The probability transducer gives probabilities that are easily combined with the set of utilities specific to the problem, to make a classification or a more general decision based on \emph{maximum expected utility}, according to the principles of decision theory. This procedure is computationally inexpensive: a low-dimensional matrix multiplication followed by an \enquote{argmax}. We have called \enquote{augmentation} the joint use of transducer and utility-maximization. The utilities employed by the augmentation can also differ from one tested item to the other, without any changes to the computational costs.

We have demonstrated the use of the probability transducer and augmentation on a \RF\ and a \CNN\ in a drug-discovery problem: classifying molecules as \enquote{inactive} or \enquote{active}. The problem has a naturally high class imbalance, and standard \ml\ classifiers often have nearly optimal performance on the dataset used to explore this problem. Yet, \emph{the augmentation led to improvements for all possible choice of utilities} underlying the classification, as shown in \fig~\ref{fig:RF_gain_UMspace}. The calculation of the two probability transducers' parameters took at most 75\,min.

The calculation of a probability transducer from a calibration dataset also provides extremely useful additional information, for free or almost so:
\begin{enumerate*}[label=(\alph*)]
\item the possible variability of the transducer function, if more calibration data were acquired; 
\item the expected utility of the whole algorithm on which the transducer is used, including the uncertainty about such utility;
\item the possibility of using the classifier in a \enquote{generative mode}, giving the probability of the output conditional on the class; this is useful when the only available data for training has different statistical properties from the real-use data.
\end{enumerate*}

\medskip

Some literature has promoted and employed the use of utilities in so called cost-sensitive learning \autocites{elkan2001,correabahnsenetal2015,lingetal2017}. One approach is to bake utilities into the loss function used at the training stage, so that the utilities can have an effect on the training of the classifier. This approach effectively wastes information and has computational disadvantages. First, since the optimal decision depends on the product of utilities and probabilities, the algorithm learns about this product only, and not about the two factors separately. \autocites[In][\sects~2--3, for example, the decision threshold of the algorithm is changed by making the algorithm  learn wrong class probabilities on purpose]{elkan2001} Yet, the utilities are known, otherwise they could not be combined with the loss function. The information about them is therefore wasted. Second, if the statistics of the data involved remain the same, but the utilities suddenly change, the classifier has to be trained anew. Such a classifier cannot be used in cases where the utilities differ from one tested item to the next (see discussion above).

The method proposed in the present work does not suffer from either of these drawbacks. The training phase needs no changes, and focuses on retrieving information about the data's statistics -- which is then extracted by the probability transducer. The full information contained in the utilities is used. And the utilities can even be changed on the fly during the use of the classifier.

\subsubsection{Future directions}
\label{sec:future}

It is possible to construct a probability transducer that takes the output from several classifiers at once. This would be the optimal way of doing \enquote{ensembling} from the point of view of the probability theory. In future work we plan to examine this possibility and compare it with standard ensembling methods.

As mentioned at the end of  \sect~\ref{sec:calibration_data}, we also plan to assess what is the best way to split available data into the training set and the calibration set, in order to have an optimal amount of mutual information between features, class, and algorithm output (training data) and a reliable transducer (calibration data).

For the demonstration of \sect~\ref{sec:demonstration} we also tried a \enquote{mixed} method: directly combining the output of the classifier (raw output for the \RF\ and standard softmax for the CNN), as it were a probability, with the utilities; and then classifying by utility maximization as usual. This method generally led to improvements with respect to the standard one, and in some cases also with respect to the probability-transducer augmentation. But on average, over the space of utility matrices, the mixed method was worse than the augmentation method, for both \RF\ and \CNN. In future work we may try to compare the performance of the two methods with different kinds of dataset.

% \mynotew{Maybe add note about how (sequential) decision theory was used during World War I; see  \textcites{good1950} around \sect~6.2 }

% \begin{figure}[t]
%   \centering
%   \includegraphics[width=\linewidth]{comparison_schema2.pdf}\\
%   \caption{Comparison scheme}
%   \label{fig:comparison scheme}
% \end{figure}

\bigskip

%%%%%%%%%%%%%%%%%%%%%%%%%%%%%%%%%%%%%%%%%%%%%%%%%%%%%%%%%%%%%%%%%%%%%%%%%%%%
%%% Acknowledgements
%%%%%%%%%%%%%%%%%%%%%%%%%%%%%%%%%%%%%%%%%%%%%%%%%%%%%%%%%%%%%%%%%%%%%%%%%%%% 
\begin{contributions}
The authors were so immersed in the development of the present work, that unfortunately they forgot to keep a detailed record of who did what.
\end{contributions}

\begin{acknowledgements}
  KD and ASL acknowledge support from the Trond Mohn Research Foundation, grant number BFS2018TMT07, and PGLPM from The Research Council of Norway, grant number 294594.

  The computations of the parameters for the probability transducer were performed on resources provided by Sigma2 -- the National Infrastructure for High Performance Computing and Data Storage in Norway (project NN8050K).

  KD would like to thank family for endless support; partner Synne for constant love, support, and encouragement; and the developers and maintainers of Python, FastAi, PyTorch, scikit-learn, NumPy and RDKit for free open source software and for making the experiments possible. 

  PGLPM thanks Maja, Mari, Miri, Emma for continuous encouragement and affection;  Buster Keaton and Saitama for filling life with awe and inspiration; and the developers and maintainers of \LaTeX, Emacs, AUC\TeX, Open Science Framework, R, Nimble, Inkscape, LibreOffice, Sci-Hub for making a free and impartial scientific exchange possible.
  % Our work was supported by the Trond Mohn Research Foundation, grant number BFS2018TMT07
%\rotatebox{15}{P}\rotatebox{5}{I}\rotatebox{-10}{P}\rotatebox{10}{\reflectbox{P}}\rotatebox{-5}{O}.
%\sourceatright{\autanet}
%\mbox{}\hfill\autanet
\end{acknowledgements}

%%%%%%%%%%%%%%%%%%%%%%%%%%%%%%%%%%%%%%%%%%%%%%%%%%%%%%%%%%%%%%%%%%%%%%%%%%%%
%%% Appendices
%%%%%%%%%%%%%%%%%%%%%%%%%%%%%%%%%%%%%%%%%%%%%%%%%%%%%%%%%%%%%%%%%%%%%%%%%%%% 

%\vspace{2\bigskipamount}

\clearpage

\renewcommand*{\appendixpagename}{Appendices: mathematical and technical details}
% \renewcommand*{\appendixname}{Appendix: test2}
% %\appendixpage
\appendix

%%%%% Kjetil's
\section{Algorithms and data used in the demonstration}
\label{sec:appendix_algorithms} 

\subsection{Data} 
  
The data comes from the open-access ChEMBL bioactivity database \autocites{bentoetal2014}. The dataset used in the present work was introduced by \textcite{koutsoukasetal2017}. The data consist in structure-activity relationships from version 20 of ChEMBL, with Carbonic Anhydrase II (ChEMBL205) as protein target.

\subsection{Pre-processing} 

For our pre-processing pipeline, we use two different methods to represent the molecule, one for the Random Forest (RF) and one for the Convolutional Neural Network (CNN). The first method turns the molecule into a hashed bit vector of circular fingerprints called Extended Connectivity Fingerprints (ECFP) \autocites{rogersetal2010}. From our numerical analysis, there was little to no improvement using a 2048-bit vector over a 1024-bit vector. 
  
For our convolutional neural network, the data is represented by converting the molecule into images of 224~pixels \texttimes\ 224~pixels. This is done by taking a molecule's SMILES (Simplified Molecular Input Line Entry System) string \autocites{davidetal2020} from the dataset and converting it into a canonical graph structure by means of RdKit \autocites{rdkit2017}. This differs from ECFP in that it represents the actual spatial and chemical structure (or something very close to it) of the molecule rather than properties generated from the molecule.

The dataset has in total 1631 active molecules and 16310 non-active molecules which act as decoys. For training, the active molecules are oversampled, as usually done with imbalanced datasets \autocites{provost2000}, to match the same number of non-active molecules.
% The data is therefore unbalanced and we solve this by giving the loss functions different weights for each class, which give the same result but have better performance as oversampling.

\subsection{Prediction}

Virtual screening is the process of assessing chemical activity in the interaction between a compound (molecule) and a target (protein). The goal of the machine learning algorithms is to find structural features or chemical properties that show that the molecule is active towards the protein \autocites{green2019}. Deep neural networks have previously been shown to outperform random forests and various linear models in virtual high-throughput screening and in quantitative structure-activity relationship (QSAR) problems \autocite{koutsoukasetal2017}.
  
\subsection{Chosen classifiers} 

The algorithms and methods used to create the models have previously been shown to give great results for a lot of different fields.

\subsubsection{\textit{Random Forest}}

The first machine learning model used in the experiments is an RF model implemented in sci-kit learn \autocites{pedregosaetal2011}. RF is an ensemble of classifying or regression trees where the majority of votes is chosen as the predicted class \autocites{breiman2001b}. It is known for being robust when dealing with a large number of features (as in our case), being resilient to over-fitting, and achieving good performance. And has already been shown to deliver powerful and accurate results in compound classification and QSAR analysis \autocites{svetniketal2003}. The following parameters were used when training the model: 

\medskip

%\fbox{\parbox{\textwidth}{
\begin{framed}
  \begin{description}
  \item[Number of trees:] 200 
  \item[Criterion:] Entropy 
  \item[Max Features:] Square root
  \end{description}
\end{framed}
%}}

\subsubsection{\textit{Convolutional Neural Network}} 
The second model is a pre-trained residual network (ResNet) \autocites{heetal2016} with 18 hidden layers trained on the well-known ImageNet dataset  \autocites{russakovskyetal2015} by using the PyTorch framework \autocites{paszkeetal2019}. ResNet has shown to outperform other pre-trained convolutional neural network models \autocites{heetal2016}. A ResNet with 34 hidden layers showed little to no performance gain, so we chose to go with the simpler model. The model is trained with the following hyperparameters:
\medskip

%\fbox{\parbox{\textwidth}{
\begin{framed}
  \begin{description}
  \item[Learning rate:] 0.003 
  \item[Optimization technique:] Stochastic Gradient Descent 
  \item[Activation Function:] Rectified linear unit (ReLU) 
  \item[Dropout:] 50\% 
  \item[Number of epochs:] 20 
  \item[Loss function:] Cross-entropy loss
  \end{description}
\end{framed}
%}}

\subsection{Dateset split}

The data set is split into four parts:
\begin{itemize}
\item Training set: 45\% of the dataset to train the model.
\item Validation set: 15\%, for validating the model after each epoch.
\item Calibration set: 20\%, for calibrating the probability transducer.
\item Demonstration set: 20\%, for evaluation.
\end{itemize}
%%%%

\bigskip

\section{Mathematical details and computation of the transducer}
\label{sec:maths_transducer}

The notation is the one used in \sect~\ref{sec:calculation_transducer}: the class is denoted $c$ and the algorithm output $y$. In our demonstration $c$ takes on values in $\set{0,1}$, and $y$ either in $\clcl{0,1}$ or in $\RR^{2}$; but the method can be applied to more general cases, such as continuous but low-dimensional spaces for both $c$ and $y$, or combinations of continuous and discrete spaces. For convenience we use a single symbol for the pair $d \defd (c, y)$.

For general references about the probability calculus and concepts and specific probability distributions see \cites{jaynes1994_r2003,mackay1995_r2005,jeffreys1939_r1983,gregory2005,bernardoetal1994_r2000,hailperin1996,good1950,fentonetal2019,johnsonetal1969_r1996,johnsonetal1969b_r2005,johnsonetal1970_r1994,johnsonetal1970b_r1995,kotzetal1972_r2000}.

\subsection{Exchangeability and expression for the probability transducer}
\label{sec:deFinetti}

There is a fundamental theorem in the probability calculus that tells us how extrapolation from known units -- molecules, patients, widgets, images -- to new, unknown units takes place: de~Finetti's theorem \autocites[\chap~4]{bernardoetal1994_r2000}{dawid2013,definetti1929,definetti1937}. It is a consequence of the assumption that our uncertainty is invariant or \enquote{exchangeable} under permutations of the labelling or order of the units (therefore it does not apply to time series, for example).

De~Finetti's theorem states that the probability density of a value $d_{0}$ for a new unit \enquote{0}, conditional on data $D$, is given by the following integral:
\begin{equation}
  \label{eq:prob_is_expe_freq}
  \p(d_{0} \| D) = \int\! F(d_{0})\ \wf(F \| D)\ \di F \ ,
\end{equation}
which can be given an intuitive interpretation. We consider every possible long-run frequency distribution $F(d)$ of data; give it a weight density $\wf(F \| D)$ which depends on the observed data; and then take the weighted sum of all such long-run frequency distributions.

The weight $\wf(F \| D)$ given to a frequency distribution $F$ is proportional to two factors:
\begin{equation}
  \label{eq:weight_F_two_factors}
  \wf(F \| D) \propto F(D)\ \wfo(F) \ .
\end{equation}
\begin{itemize}
  \item The first factor (\enquote{likelihood}) $F(D)$ quantifies how well $F$ \emph{fits} known data of the same kind, in our case the calibration data $D \defd \set{d_{1}, \dotsc, d_{M}}$. It is simply proportional to how frequent the known data would be, according to $F$:
  \begin{equation}
    \label{eq:factor_likelihood}
    F(D) \defd F(d_{1})\cdot F(d_{2})\cdot \dotsm\cdot F(d_{M})
    \equiv \exp\biggl[M \sum_{d} \hat{F}(d) \ln F(d)\biggr] \ ,
  \end{equation}
where $\hat{F}(d)$ is the frequency distribution observed in the data.
  
\item The second factor (\enquote{prior}) $\wfo(F)$ quantifies how well $F$ \emph{generalizes} beyond the data we have seen, owing to reasons such as physical or biological constraints for example. In our case we expect $F$ to be somewhat smooth in $X$ when this variable is continuous \autocites[Cf.][]{goodetal1971}. No assumptions are made about $F$ when $X$ is discrete.
  %%\mynotew{add plot of prior transducer curves?}
\end{itemize}
Formula~\eqref{eq:weight_F_two_factors} is just Bayes's theorem. Its normalization factor is the integral $\int F(D)\, \wfo(F)\, \di F$, which ensures that $\wf(F)$ is normalized.

The exponential expression in \eqn~\eqref{eq:factor_likelihood} is proportional to the number $M$ of data, and in it we recognize the cross-entropy between the observed frequency distribution $\hat{F}$ and $F$. This has two consequences. First, it makes the final probability $p(d_{0})$ increasingly identical with the distribution $\hat{F}(d)$ observed in the data, because the average~\eqref{eq:prob_is_expe_freq} gets more and more concentrated around $\hat{F}$. Second, a large amount of data indicating a non-smooth distribution $F$ will override any smoothness preferences embodied in the second factor. Note that no assumptions about the shape of $F$ -- Gaussians, logistic curves, sigmoids, or similar -- are made in this approach (compare \fig~\ref{fig:prob_curve_RF_samples}).

\subsection{Conditional probabilities}
\label{sec:cond_probs}

From \eqn~\eqref{eq:prob_is_expe_freq}, which is expressed in terms of joint probabilities for $c_{0}$ and $y_{0}$, there are two ways of obtaining the probability of $c_{0}$ conditional on $y_{0}$, which we report without proof:
\begin{description}
\item[Exchangeable output:] if the newly observed value $y_{0}$ of the output is considered to be exchangeable with (or representative of) the $y$ values in the calibration data, then
  \begin{equation}
    \label{eq:c_cond_y_exchangeable}
    \p(c_{0} \| y_{0}, D, \texts{exch.}) =
    \frac{\p(c_{0}, y_{0} \| D)}{ \p(y_{0} \| D)} =
    \frac{\int\! F(c_{0}, y_{0})\ \wf(F \| D)\ \di F}{
      \int\! F(y_{0})\ \wf(F \| D)\ \di F} \ ,
  \end{equation}
  where $F(y_{0}) = \sum_{c}F(c, y_{0})$. This expression is effectively doing two things: first, implicitly updating the probability distribution for $y$, taking as new evidence the observed $y_{0}$; second, yielding the conditional distribution of $c_{0}$ given $y_{0}$.
    \item[Non-exchangeable output:] if the newly observed value $y_{0}$ of the output is \emph{not} considered to be exchangeable with the $y$ values in the calibration data, then
  \begin{equation}
    \label{eq:c_cond_y_exchangeable}
    \p(c_{0} \| y_{0}, D, \texts{non-exch.}) =
    \int\! \frac{F(c_{0}, y_{0})}{F(y_{0})}\ \wf(F \| D)\ \di F \ ;
  \end{equation}
  This expression does not implicitly update of the distribution for $y$.
\end{description}

The second formula should be used if new data are considered to lead to a  distribution of features different from that of the calibration data -- although the basic assumption that the \emph{conditional} distributions of classes given features are the same still holds \autocites[see][for a thorough discussion of these two cases]{lindleyetal1981}.

The two formulae converge to one another and to the long-run conditional probability of $c$ given $y$, as the number of calibration data increases. With a large number of calibration data, say hundreds or more, the values obtained from the two formulae are negligible.

\subsection{Representation of the long-run distribution and Markov-chain Monte Carlo sampling}
\label{sec:MCMC}

The integral in~\eqref{eq:prob_is_expe_freq} is calculated in either of two ways, depending on whether $d$ is discrete or continuous. For $d$ discrete, the integral is over $\RR^{n}$, where $n$ is the number of possible values of $d$, and can be done analytically. For $d$ with continuous components, the integral is numerically approximated by a sum over $T$ representative samples, obtained by Markov-chain Monte Carlo, of distributions $F$ according to the weights~\eqref{eq:weight_F_two_factors}:
\begin{equation}
  \label{eq:MC_approx}
  \p(d_{0} \| D) = \int\! F(d_{0})\ \wf(F \| D)\ \di F
  \approx
\frac{1}{T}\sum_{t=1}^{T} F_{t}(d_{0}) \ .  
\end{equation}
The error of this approximation can be calculated and made as small as required by increasing the number of Monte Carlo samples.

We must find a way to express any kind of distribution $F(d)$. As mentioned in \sect~\ref{sec:calculation_transducer}, this is done by writing it as
\begin{equation}
  \label{eq:representation_F}
  F(c, y) = \sum_{l} w_{l}\ A(c \| \alpha_{l})\ B(y \| \beta_{l}) \ ,
\end{equation}
where the sum has a large number of terms \autocites[see][on why the number of terms does not need to be infinite]{ishwaranetal2002b}, $\set{w_{l}}$ are normalized weights, and $A(c\|\alpha)$, $B(y \| \beta)$ are distributions, possibly the product of further one-dimensional distributions. Effectively we are expressing $F(\dotv)$ by the \enquote{coordinates} $(w_{l}, \alpha_{l}, \beta_{l})$ in a space of extremely high dimensions.
%\mynotew{add remark about number of clusters and ref...}

This representation \autocites[promoted by][]{dunsonetal2011}[see also][]{rasmussen1999} has several advantages:
\begin{itemize}
\item Its marginal distributions for $c$ and $y$ are also of the form~\eqref{eq:representation_F}, as shown in \sect~\ref{sec:calculation_transducer}, and easily computable.
\item Its conditional distributions for $c$ given $y$ and vice versa have also a form similar to \eqn~\eqref{eq:representation_F} and easily computable.
\item It can be used with conjugate priors.
\item The final probability $p(c,y)$, approximated by the sum~\eqref{eq:MC_approx}, has also the form~\eqref{eq:representation_F}:
  \begin{equation}
    \label{eq:sum_of_sums_for_prob}
    \p(c_{0}, y_{0}) \approx
    \sum_{t,l} \frac{w_{t,l}}{T}
    \ A(c_{0} \| \alpha_{t,l})\ B(y_{0} \| \beta_{t,l}) \ .
  \end{equation}
This is the expression given in \sect~\ref{sec:calculation_transducer} with $k$ running over both indexes $t,l$ and with $q_{k} \defd w_{t,l}/T$.
\end{itemize}

In the demonstration of \sect~\ref{sec:demonstration}, where the class variable $c$ takes on conventional values $\set{0,1}$, we use a Bernoulli distribution for the class variable $c$:
\begin{equation}
  \label{eq:distr_class}
  A(c \| \alpha) = c\ \alpha + (1-c)\ (1-\alpha) \equiv
  \begin{dcases*}
    \alpha& if $c=1$, \\ 1-\alpha & if $c=0$ \ ,
  \end{dcases*}
\end{equation}
both in the case of the \RF\ and of the \CNN. For the output variable we use
a Gaussian distribution in the case of the \RF, $\beta \equiv (\mu,\sigma)$ being its mean and standard deviation:
\begin{equation}
  \label{eq:distr_output}
  B(y \| \beta) = \No(y \| \mu, \sigma) \defd \frac{1}{\sqrt{2\pu\sigma^{2}}} \exp\biggl[-\frac{(y-\mu)^{2}}{2\sigma^{2}}\biggr] \ .
\end{equation}
The \rf\ output is actually bounded, $y\in \clcl{0,1}$, and this Gaussian should in principle be truncated; we did not use any truncation as the error committed is small and the computation much faster. In the case of the  \CNN's output we use a product of two Gaussians, each with its own parameters.

In both cases, the sum of the representation~\eqref{eq:representation_F} has 64 terms, and the approximating sum~\eqref{eq:MC_approx} 4096 terms.

\medskip

The samples $\set{F_{t}(\dotv)}$ of the sum~\eqref{eq:MC_approx} -- these are samples of the distribution $\wf(F)$ -- are obtained through Markov-chain Monte Carlo; specifically Gibbs sampling \autocites{neal1993}[\chap~29]{mackay1995_r2005}.
Effectively we obtain samples of the coordinates $(w_{l}, \alpha_{l}, \beta_{l})$, and the prior $\wfo(F)$ is a prior over these coordinates.

For the demonstration of \sect~\ref{sec:demonstration} we use a Dirichlet distribution for $(w_{l})$, a beta distribution for $(\alpha_{l})$, a Gaussian distribution for $(\mu_{l})$, and a gamma distribution for $(1/{\sigma_{l}}^{2})$.

The Markov-chain Monte Carlo sampling scheme is implemented using the R \autocites{rcoreteam1995} package \textbf{nimble} \autocites{nimble2016_r2021}, and uses 16 parallel chains \autocites[\addcomma\ scripts \texttt{RFmcmc.R} and \texttt{NNmcmc.R}]{dyrlandetal2022c}. Alternatively one can use the R package \textbf{prova} \autocites{portamana2024} for automated Markov-chain Monte Carlo. The sampling took approximately 45\,min for the \RF\ and 75\,min for the \CNN, wall-clock time. The resulting parameters are available in our supplementary data\footcites[\addcomma\ files \texttt{transducer\_params-Random\_Forest.zip} and \texttt{transducer\_params-Neural\_Net.zip}]{dyrlandetal2022c}.

\subsection{Assessment of the possible variability of the probability}
\label{sec:variability_prob}

In \sect~\ref{sec:variability_curve} we mentioned that the calculation of the transducer parameters automatically also tells us how much the probabilities curves could change if we used more data for the calibration. The range of this possible variability was shown for example in \figs~\ref{fig:prob_curve_RF}, \ref{fig:prob_curve_RF_inverse}, and \ref{fig:prob_curve_CNN_section} of the demonstration.

This possible variability is encoded in the weight $\wf(F \| D)$, which can be interpreted as the probability distribution of the long-run frequency distribution $F$; remember that the probability $\p(d_{0} \| D)$, for the new unit, \eqn~\eqref{eq:prob_is_expe_freq}, becomes closer and closer to the distribution $F$ at which $\wf(F \| D)$ peaks, as the number of data increases.

\begin{figure}[t]
  \centering
  \includegraphics[width=0.75\linewidth]{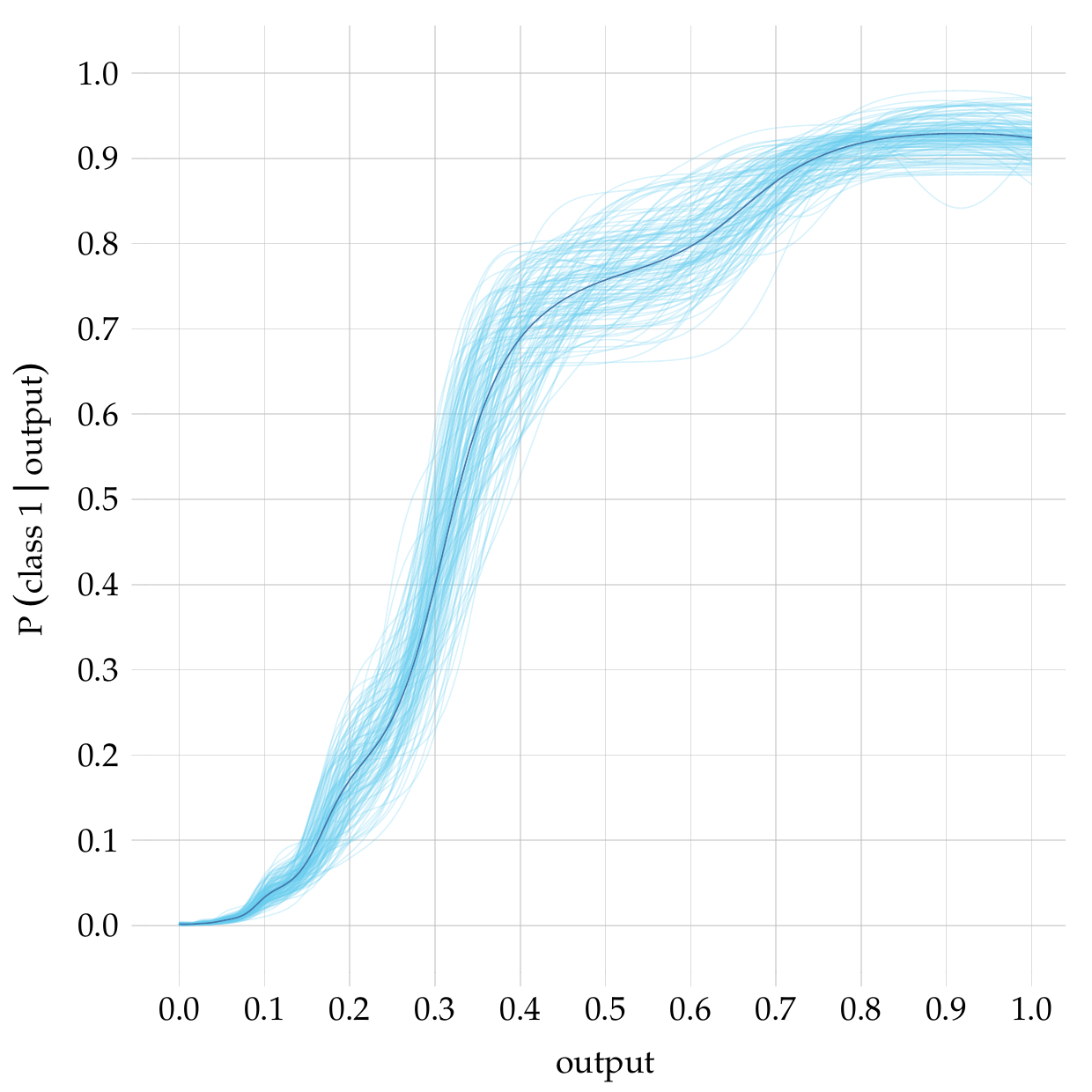}\\
  \caption{Samples (\textcolor{cyan}{thin light-blue curves}) of probable transducer curves (for class~1) that could be obtained if we had more calibration data in the demonstration of \sect~\ref{sec:demonstration}. They are samples obtained from the distribution $\wf(F \| D)$ of \eqn~\eqref{eq:weight_F_two_factors}. According to the probability calculus, the curve to be used for a new unit is their average (\textcolor{blue}{thicker dark-blue curve}), which is the same as plotted in \fig~\ref{fig:prob_curve_RF}.}
  \label{fig:prob_curve_RF_samples}
\end{figure}

In the approximation~\eqref{eq:MC_approx}, the probability $\wf(F \| D)$ is effectively represented by a large number of samples $\set{F_{t}}$ from it. Plotting these samples alongside $\p(d_{0} \| D)$ gives an approximate idea of how the latter probability could change with new data. Figure~\ref{fig:prob_curve_RF_samples} shows a small number of such samples for the transducer curve of the \RF\ of \sect~\ref{sec:demonstration} (\cf\ \fig~\ref{fig:prob_curve_RF}). For fixed $d$, the samples $\set{F_{t}(d)}$ also give estimates of the quantiles of such change. This is how the ranges of \figs~\ref{fig:prob_curve_RF}, \ref{fig:prob_curve_RF_inverse}, \ref{fig:prob_curve_CNN_section} were obtained.

An analogous discussion holds for the marginal and conditional probabilities that we can obtain from $\p(d_{0} \| D)$.

\subsection{Assessment of the augmented algorithm's long-run utility yield}
\label{sec:algorithm_yield}

Besides making a decision -- such as choosing a class -- for each new unit, we generally must also decide which algorithm to use for such a future task, among a set of candidates. This latter decision depends on the future performance of each algorithm, which in turn depends on the decision that the algorithm will make for each new unit. Two kinds of unknown accompany this double decision: we do not know the classes of the future units, and we do not know which outputs each algorithm will give for the future data.

This more complex kind of decision \amp\ uncertainty problems are also dealt with decision theory. Their theory is presented  and applied step-by-step in the humorous lectures by \cites[\chap~2]{raiffa1968_r1970}; other references are  \autocites{lindley1971_r1988}[\sect~2.2]{bernardoetal1994_r2000}{prattetal1995_r1996,raiffaetal1961_r2000,luceetal1957}. We here give only a sketch and refer to the works above for details.

\begin{figure}[!t]
  \centering
  \includegraphics[width=\linewidth]{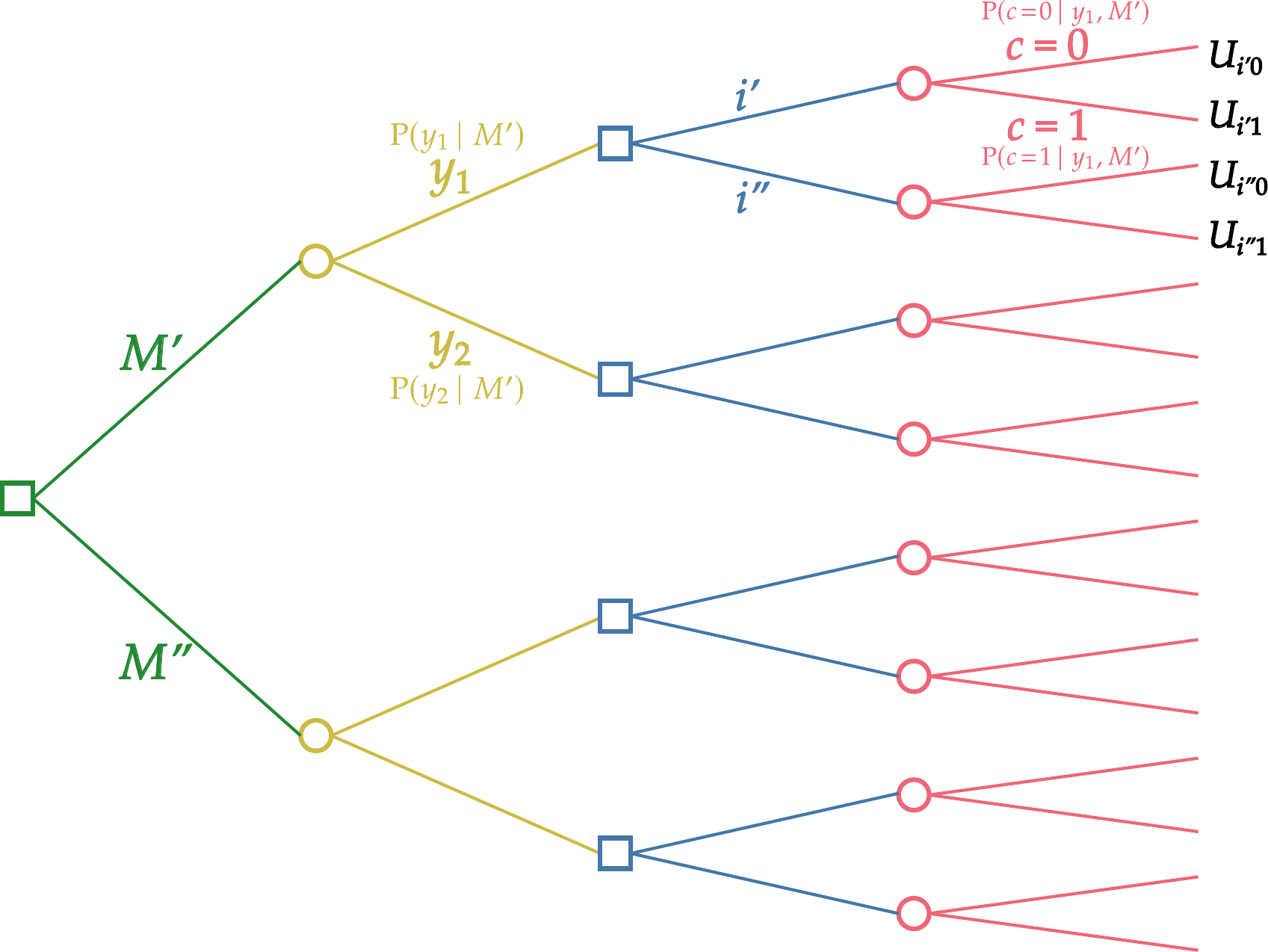}\\
  \caption{Decision tree for the choice of algorithm. From left to right: Decision node about the algorithm (green), uncertainty node about the algorithm's output (yellow), decision node (\eg\ about class) for the inspected item (blue), uncertainty node about the class (red). Only some example nodes and branches are labelled.}
  \label{fig:decision_tree}
\end{figure}

Our double decision \amp\ uncertainty problem can be represented as a \emph{decision tree}. A very simplified example is illustrated in \fig~\ref{fig:decision_tree}.

We imagine to have to choose between two classification algorithms $M'$ and $M''$. This choice corresponds to the \emph{decision node} on the left (green). Decision nodes are represented by squares.

If we choose to use algorithm $M'$, then it may happen that it will give either output $y_{1}$ or $y_{2}$ when applied to a new unit. We are uncertain about which output will occur, with probabilities $\P(y_{1} \| M')$ and $\P(y_{2} \| M')$. This uncertainty corresponds to an \emph{uncertainty node} (yellow). Uncertainty nodes are represented by circles.

Once the output of the algorithm is known, we must decide (blue decision nodes) among choices $i'$ and $i''$, for example to choose whether to consider the new unit as class~0 or class~1.

The unit will turn out to be class~0 or class~1: we are uncertain about which (red decision nodes), with probabilities $\P(c\mo0 \| y_{1}, M')$, $\P(c\mo1 \| y_{1}, M')$ if the output was $y_{1}$, and with probabilities $\P(c\mo0 \| y_{2}, M')$, $\P(c\mo1 \| y_{2}, M')$ if the output was $y_{2}$.

Finally, depending on our choice between $i'$ and $i''$ and on the actual class, we will gain one of the four utility amounts $U_{i1\, 0}$, $U_{i'\, 1}$, $U_{i''\, 0}$, $U_{i''\, 1}$. These are the elements of the utility matrix discussed in \sect~\ref{sec:utility_classification}; we have seen concrete numerical examples in the demonstration of \sect~\ref{sec:demonstration}.

An analogous analysis and probabilities hold if we choose algorithm $M''$ (the output space of this algorithm can be different from that of $M'$).

The basic procedure of this decision problem is to first calculate expected utilities starting from the terminal uncertainty nodes, making optimal decisions at the immediately preceding decision nodes. Each such decision will therefore have an associated utility equal to its corresponding maximal expected utility. The same procedure is then applied to the uncertainty nodes about the outputs. In this way each algorithm receives a final expected utility; in formulae,
\begin{equation}
  \label{eq:algorithm_utility}
\texts{utility of algorithm $M$} =  \sum_{y} \biggl[\max_{i}\set[\bigg]{\sum_{c} U_{ic}\ \P(c \| y, M)}\biggr]\
  \P(y \| M) \ .
\end{equation}
Either sum is replaced by an integral over a density if the related quantity, $y$ or $c$, is continuous.

What is important in the formula above is that the probabilities for the outputs and the conditional probabilities for the classes given the outputs are known: \emph{they are the ones calculated for the transducer from the calibration set}.

The expected utilities of table~\ref{tab:algorithms_utilities}, \sect~\ref{sec:yield_of_classifier}, were calculated with the formula above (the integral over $y$ being approximated by a sum over a dense grid).

\medskip

These values are \emph{expected} utilities, though. One may ask: what is the probability that the final utility of one model will actually be higher or lower than the other's?

We can answer this question, again thanks to de~Finetti's formula~\eqref{eq:prob_is_expe_freq}, similarly to how we did with the variability of the transducer curves, explained in the previous \sect~\ref{sec:variability_prob}. The long-run utility of the algorithm $M$ is given by formula~\eqref{eq:algorithm_utility} but with the probabilities replaced by the long-term frequencies $F(c\|y)$ and $F(y)$. The probability of this long-run utility is then determined by the density $\wf(F)$ represented by a set of samples. Calculating the long-run utility for each sample we can finally construct a probability histogram for each algorithm's utility. The histograms of \fig~\ref{fig:algorithms_utilities_histograms} are obtained this way. From them we can also calculate the probability that an algorithm's utility $u'$ will be higher than another's utility $u''$, corresponding to the integral
\begin{equation}
  \label{eq:compare_alg_utilities}
\iint \delt(u' > u'')\ \p(u')\ \p(u'')\ \di u'\ \di u'' \ .
\end{equation}

%%%% examples use empheq
%   \begin{empheq}[left={\mathllap{\begin{aligned}    \de\yF_{\yc}/\de\yp&=0\text{:} \\
%         \de\yF_{\yc}/\de\ym&=0\text{:}\\ \de\yF_{\yc}/\de\yl&=0\text{:}\end{aligned}}\qquad}\empheqlbrace]{align}
%     \label{eq:con_p}
% %    \de\yF_{\yc}/\de\yp &\equiv
%     -\ln\yp + \ln\yq + \yl\yM + \ym\yu &=0,\\
%     \label{eq:con_u}
% %    \de\yF_{\yc}/\de\ym &\equiv
%     \yu\yp-1 &=0,\\
%     \label{eq:con_l}
%     %\de\yF_{\yc}/\de\yl &\equiv
%     \yM\yp-\yc &=0.
%   \end{empheq}
%%%%
% \begin{empheq}[box=\widefbox]{equation}
%   \label{eq:maxent_question}
%   \p\bigl[\yE{N+1}{k} \bigcond \tsum\yo\yf{N}\in\yA, \yM\bigr] = \mathord{?}
% \end{empheq}

%%\setlength{\intextsep}{0ex}% with wrapfigure
%%\setlength{\columnsep}{0ex}% with wrapfigure
%\begin{figure}[p!]% with figure
%\begin{wrapfigure}{r}{0.4\linewidth} % with wrapfigure
%  \centering\includegraphics[trim={12ex 0 18ex 0},clip,width=\linewidth]{maxent_saddle.png}\\
%\caption{caption}\label{fig:comparison_a5}
%\end{figure}% exp_family_maxent.nb

\clearpage

%%%%%%%%%%%%%%%%%%%%%%%%%%%%%%%%%%%%%%%%%%%%%%%%%%%%%%%%%%%%%%%%%%%%%%%%%%%%
%%% Bibliography
%%%%%%%%%%%%%%%%%%%%%%%%%%%%%%%%%%%%%%%%%%%%%%%%%%%%%%%%%%%%%%%%%%%%%%%%%%%% 
\renewcommand*{\finalnamedelim}{\addcomma\space}
\defbibnote{prenote}{{\footnotesize (\enquote{de $X$} is listed under D,
    \enquote{van $X$} under V, and so on, regardless of national
    conventions.)\par}}
% \defbibnote{postnote}{\par\medskip\noindent{\footnotesize% Note:
%     \arxivp \mparcp \philscip \biorxivp}}

\printbibliography[prenote=prenote%,postnote=postnote
]

\end{document}